\documentclass{article}

\PassOptionsToPackage{numbers}{natbib}
\usepackage[final]{neurips_2025}
\usepackage{amsmath}

\usepackage{amsfonts}
\usepackage{amssymb}
\usepackage[most]{tcolorbox}
\usepackage{enumitem}

\usepackage[utf8]{inputenc} 
\usepackage[T1]{fontenc}    
\usepackage{hyperref}       
\usepackage{url}            
\usepackage{booktabs}       
\usepackage{amsfonts}       
\usepackage{nicefrac}       
\usepackage{microtype}      
\usepackage[table,xcdraw]{xcolor}
\usepackage{multirow}
\usepackage{color}
\usepackage{graphicx}
\usepackage{wrapfig}
\usepackage{subcaption}
\usepackage{courier}
\usepackage[dvipsnames]{xcolor}
\usepackage{lscape}
\usepackage{capt-of}
\usepackage{minitoc}
\usepackage{titletoc}

\newcommand{\props}{\texttt{ProPS}}
\newcommand{\propss}{\texttt{ProPS$^+$}}
\newcommand{\promptvariable}{\textcolor{Variable}}
\definecolor{Variable}{HTML}{AF7841}
\definecolor{tbox_bg}{HTML}{ebfbff} 
\definecolor{tbox_frame}{HTML}{00a1d7} 
\definecolor{title_color}{HTML}{c4c4c4}
\definecolor{best_result}{HTML}{CCf2FF}

\title{Prompted Policy Search:  Reinforcement Learning through Linguistic and Numerical Reasoning in LLMs}

\author{%
Yifan Zhou$^{1,}$\thanks{Corresponding Authors.} ,  Sachin Grover$^{1,*}$, Mohamed El Mistiri$^{1}$, Kamalesh Kalirathnam$^1$, \\
\textbf{Pratyush Kerhalkar$^1$, Swaroop Mishra$^{3, }$\thanks{Currently working at Microsoft AI} , Neelesh Kumar$^2$, Sanket Gaurav$^2$,}\\
\textbf{Oya Aran$^2$, Heni Ben Amor$^1$} \\
$^1$Interactive Robotics Lab, Arizona State University \\
$^2$Research \& Development, Procter \& Gamble\\
$^1$\texttt{\{yzhou298,sgrover6,melmisti,kamales1,pkerhalk,hbenamor\}@asu.edu} \\
$^2$\texttt{\{kumar.n.40,gaurav.s,aran.o\}@pg.com},
$^3$\texttt{swaroopranjanmishra@gmail.com}
}

\begin{document}

\maketitle

\begin{abstract}
Reinforcement Learning (RL) traditionally relies on scalar reward signals, limiting its ability to leverage the rich semantic knowledge often available in real-world tasks. In contrast, humans learn efficiently by combining numerical feedback with language, prior knowledge, and common sense. We introduce Prompted Policy Search (\props), a novel RL method that unifies numerical and linguistic reasoning within a single framework. Unlike prior work that augment existing RL components with language, \props\ places a large language model (LLM) at the center of the policy optimization loop—directly proposing policy updates based on both reward feedback and natural language input. We show that LLMs can perform numerical optimization in-context, and that incorporating semantic signals, such as goals, domain knowledge, and strategy hints can lead to more informed exploration and sample-efficient learning. \props\ is evaluated across 15 Gymnasium tasks, spanning classic control, Atari games, and MuJoCo environments, and compared to seven widely-adopted RL algorithms (e.g., PPO, SAC, TRPO). It outperforms all baselines on 8 out of 15 tasks and demonstrates substantial gains when provided with domain knowledge. These results highlight the potential of unifying semantics and numerics for transparent, generalizable, and human-aligned RL.

\end{abstract}

\section{Introduction}

Reinforcement Learning (RL)~\citep{sutton1998reinforcement} represents a foundational paradigm shift within the broader field of machine learning. It allows  autonomous agents to learn optimal behaviors through interactions with their environment, i.e., through repeated trial and error. Over the past decades, RL has resulted in remarkable successes across a range of challenging domains, including mastering strategic games such as Backgammon~\citep{tesauro1994td} and Go~\citep{silver2017mastering}, achieving human-level performance in robot table-tennis~\citep{d2024achieving} or contributing to scientific breakthroughs such as protein folding~\citep{jumper2021highly}. 
Traditional RL relies exclusively on numerical feedback in the form of scalar rewards. By contrast, humans often learn and reason using natural language, prior knowledge, and common sense \citep{lupyan2016language, nefdt2020role}. Many real-world tasks are accompanied by rich linguistic context such as manuals, domain descriptions, and expert instructions which standard RL algorithms are unable to exploit. Yet, this information can serve as a powerful inductive bias: guiding exploration, encoding constraints, and expressing useful heuristics to accelerate learning.

To bridge this gap between numerics and semantics, we introduce Prompted Policy Search (\props), a new method that unifies numerical and linguistic reasoning within a single framework. \props\ enables language models to process and act on both reward signals and natural language inputs, such as high-level goals, domain knowledge, or strategic hints. This results in a more informed and adaptable policy search process. While prior works have used Large Language Models (LLMs)~\cite{minaee2025largelanguagemodelssurvey} to augment specific components of the RL pipeline (e.g., reward shaping~\citep{yu2023language}, Q-function modeling~\citep{Wu2024EnhancingQW}, or action generation~\citep{han2024}), these approaches still depend on conventional RL algorithms for optimization. In contrast, we show that LLMs can directly perform policy search, treating optimization as an in-context reasoning problem. To this end, first, we demonstrate that LLMs are capable of numerical optimization for RL tasks. We then extend this capability to incorporate linguistic abstractions, enabling a unified reasoning strategy where semantic and quantitative signals complement one another. The resulting approach accelerates convergence by incorporating prior knowledge, enforcing constraints, and refining exploration. Moreover, it offers additional transparency by providing natural language justifications of the proposed policy updates: an essential feature for domains requiring transparency, safety, and human oversight. Our primary contributions are as follows:
\begin{itemize}
    \item[($C_1$)] Unifying Numerical and Linguistic Reasoning for RL: We propose Prompted Policy Search (\props), a framework to integrate scalar reward signals with natural language guidance in a unified optimization loop, enabling language models to reason over both quantitative feedback and semantic abstractions.

    \item[($C_2$)] Flexible Integration of Human-Centric Knowledge: \props\ leverages the symbolic and generalization capabilities of LLMs to incorporate domain knowledge, goals, and heuristics via prompts thereby enabling more transparent, and sample-efficient learning in RL tasks.

    \item[($C_3$)] LLMs as In-Context Policy Optimizers: We show that pretrained large language models can perform policy search directly, using only reward feedback without relying on external optimizers or pretraining on RL-specific data. We also show that small, lightweight LLMs can be fine-tuned for better policy search performance.

    \item[($C_4$)] Comprehensive Empirical Validation Across Benchmarks: We evaluate ProPS on 15 diverse RL tasks across Gymnasium environments, comparing it against seven standard RL algorithms. ProPS achieves state-of-the-art results in more than half of the tasks, and shows measurable improvements when incorporating expert language guidance.

\end{itemize}

\section{Related Works}
A growing body of research is exploring the integration of natural language into RL pipelines. Early work in this space used language to augment policy learning or enhance interpretability~\citep{pmlr-v155-goyal21a, AAAI04ws-pillar, narasimhan18}, but such efforts were often limited to synthetic inputs or constrained corpora~\citep{ijcai2019p880}, reducing their applicability to real-world settings. Large language models (LLMs) have reignited interest in combining language with RL, enabling free-form natural language to inform learning in more complex domains~\citep{yan2025efficient, zhou24}. Most existing LLM+RL approaches incorporate language by modifying specific components of the RL pipeline. A common direction involves using LLMs to generate reward functions based on human  instructions~\citep{kwon2023reward, yao2024anybipe,yu2023language}. For instance, DrEureka~\citep{ma2024dreureka} constructs language-driven reward functions and domain randomization strategies to facilitate sim-to-real transfer, though it ultimately relies on conventional optimizers like PPO for training. Other approaches translate multi-turn instructions into executable reward code for use by downstream control algorithms such as model-predictive control~\citep{Liang2024}. However, a fundamental limitation of reward-based integration lies in the difficulty of mapping high-dimensional, often ambiguous language into scalar reward values. This mismatch can lead to biased or brittle reward functions, potentially incentivizing unintended behaviors~\citep{sinan2024}. Beyond reward modeling, LLMs have been used to suggest or refine actions directly. For example,~\citet{han2024} leverage domain knowledge from FAA flight manuals to guide action selection in flight control, while~\citet{Wu2024EnhancingQW} employ LLMs to modulate entries in a Q-table. Other works dynamically decompose tasks into subtasks and choose between solving them with language models or traditional RL agents~\citep{liu2024rlgpt}. In contrast to these approaches, our work places the LLM at the center of the learning process itself. Rather than serving as an auxiliary component (e.g., for reward shaping or action selection), the LLM performs the policy search by directly proposing parameter updates. Importantly, our approach differs from methods that treat the LLM as the policy (e.g., directly outputting actions or plans~\citep{brohan2023can}). In \props, the LLM is responsible for discovering and refining policy parameters, but not for producing actions at inference time. Our work is closest in spirit to the work on in-context RL in~\citep{laskin2023incontext, zisman2024emergence}, though prior efforts in that direction aim to distill existing numerical RL methods into Transformers that are trained from scratch. The work in~\cite{monea2025llmsincontextbanditreinforcement} focuses on in-context RL in bandit settings and restricts the approach to label classification tasks and binary rewards leaving open ``the question of applicability to more complex RL problems, where rewards are more nuanced, or where interactions comprise multiple steps''. Our work draws inspiration from OPRO -- a prompt optimization framework introduced in~\citep{yang2024large}. In OPRO, at each optimization step, a meta-prompt is used in order to generate $N$ new candidate solutions. In turn, these solutions are evaluated and ranked outside of the LLM. This strategy can be seen as a form of hill-climbing, i.e., multiple next candidates are generated by an LLM (exploration) and a limited number sub-selected outside of the LLM for further improvement (exploitation). Our work addresses a related but different challenge, namely the combination of numerical and linguistic reasoning for optimization in RL. We show that optimization can be performed entirely inside the LLM, without the need for external ranking, sub-selection or other optimization components. We also show that pretrained LLMs already possess the reasoning capabilities necessary to perform policy search, and that these capabilities can be enhanced through the fusion of semantic and numerical feedback.

\section{Methodology}
In this section, we present the description of our proposed approach and its two variants. We begin in Sec.~\ref{ref:problemstatement} by formalizing the policy search problem and our notation. Sec.~\ref{ref:prompted_overview} outlines the high-level methodology, whereas Sections~\ref{ref:props} and~\ref{ref:propss} detail two specific prompting strategies: one that relies purely on numerical feedback, and another that combines numeric and linguistic information.

\subsection{Problem Statement\label{ref:problemstatement}}

\begin{wrapfigure}{r}{0.25\textwidth}
  \vspace{-10pt}
  \centering
  \includegraphics[width=0.25\textwidth]{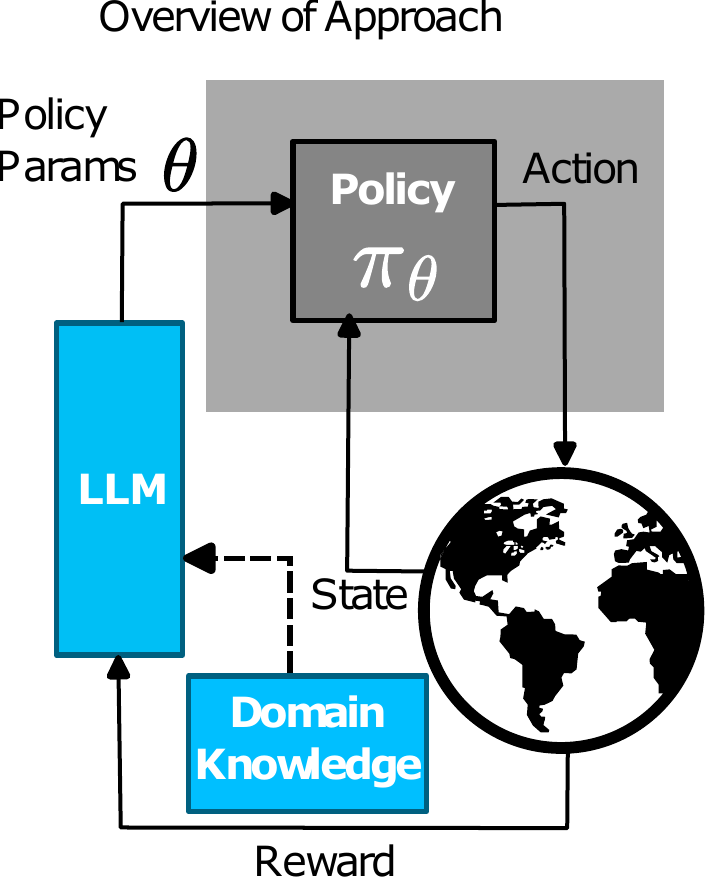}
  \caption{Overview of the approach used in \props, showing interactions between the environment, LLM, and RL.
  }
  \label{fig:overview}
  \vspace{-10pt}
\end{wrapfigure}

Policy search methods optimize a parameterized policy $\pi_\theta$ directly, without relying on an explicit value function~\citep{ROB-021}. Given parameters $\theta \in \mathbb{R}^D$, the goal is to maximize expected return $J(\theta) = \mathbb{E}[R(\boldsymbol{\tau})]$, where $\boldsymbol{\tau}$ is a trajectory of states and actions generated by $\pi_\theta$ over a finite horizon. We consider stochastic policies $\pi_\theta(\boldsymbol{a}_t|\boldsymbol{s}_t, t)$ with action $\boldsymbol{a}_t$, state $\boldsymbol{s}_t$ and timestep $t$ and the \emph{episodic setting} where a single cumulative reward $R(\boldsymbol{\tau})$ is returned at the end of each rollout. Optimization techniques include gradient ascent~\citep{sutton1999policy}, information-theoretic updates~\citep{10.5555/2898607.2898863}, and expectation-maximization~\citep{NIPS2008_7647966b}. For example, a policy gradient update follows $\theta \leftarrow \theta + \alpha \nabla\theta J(\theta)$, where $\alpha$ is the learning rate. In this work, we focus on two types of parameterized policies: \textbf{Continuous State-Space Policies}: For tasks characterized by continuous state spaces, linear policies are utilized. These are broadly represented by $\pi_\theta(\boldsymbol{s}) = \theta^T\phi(\boldsymbol{s})$, where $\phi$ is a linear feature function mapping the states into feature representation. A variety of different feature functions can be used, such as radial basis functions. For the remainder of this paper, we use a simple identity mapping for $\phi$. \textbf{Discrete State-Space Policies}: In discrete tasks, we adopt tabular policies $\pi_\theta(s) = \theta_s$, where, $\theta$ is a vector of length $|\mathcal{S}|$ (the number of states), and $\theta_s \in \mathcal{A}$ (the action space) specifies the action to take in state $s$.

\subsection{Prompted Policy Search\label{ref:prompted_overview}}

We present a novel reinforcement learning (RL) approach in which a large language model (LLM) directly generates policy parameters without relying on a conventional RL optimizer or any external optimization component beyond the reward signal. Traditional RL methods focus on numerical information (e.g., gradients with respect to the reward) and as a result cannot incorporate important task-specific knowledge that is difficult to express in numbers, such as domain semantics or user-provided guidance. To address this limitation, we introduce \textbf{Prompted Policy Search} (\props), a new method that combines \textbf{numerical reasoning} with \textbf{linguistic reasoning} to enable more flexible and informed learning. By linguistic reasoning, we mean the ability of LLMs to understand, process, and analyze natural language in order to draw (deductive and inductive) inferences and make informed decisions. This allows us to embed valuable information like prior domain knowledge, goals, or user-provided policy hints directly into the learning process using natural language. For example, traditional RL methods treat all input features as raw numbers and do not distinguish between features expressed in different units, such as meters versus kilometers. In contrast, an LLM can interpret text-based task descriptions that explain the nature and context of each feature.

Fig.~\ref{fig:overview} illustrates our approach. The LLM generates an initial parameter vector $\theta$, which specifies a policy $\pi_\theta$. This policy is then executed in the environment, and the resulting episodic reward is returned to the LLM along with a history of all previous parameters and rewards. Using this feedback, the LLM is prompted again to generate an improved version of $\theta$ with the aim of increasing expected reward. This loop continues iteratively: the LLM reasons over past performance and proposes new parameters that are likely to improve outcomes. Please note that the LLM does not participate in real-time action generation. Once policy parameters are generated, the policy operates independently: the LLM is only involved in parameter updates based on feedback. More formally, the update step in \props\ can be expressed as:
\begin{equation}
    \theta \leftarrow \mathrm{LLM}(\Gamma, \mathcal{P}) \label{eq:llm_rl}
\end{equation}
where $\Gamma = \left[\theta_{1:N}, R_{1:N}\right]$ is the history of all previous parameters $\theta_i$ suggested by the LLM along with the corresponding episodic reward $R_i$, and $N$ is the number of iterations executed so far. $\mathcal{P}$ is the prompt provided to the LLM represented as a language embedding in a latent space~\citep{tao2024llmseffectiveembeddingmodels}. Please note that updated policy parameters are generated by prompting an LLM and do not require any additional (external) computation. Subsequently, we discuss how this optimization loop can be implemented through structured prompting and introduce two different variants of our approach, namely \props\ and \propss. 

 \subsection{\props: Policy Search through Numerical Optimization with LLMs\label{ref:props}}
We begin by demonstrating how \props\ can emulate a conventional policy search setup, where learning is driven exclusively by numerical feedback from the reward function. Recent studies, in particular the work in~\cite{yang2024large} have posited that LLMs are capable of mathematical optimization through iterative prompting~\cite{amor2025saspromptlargelanguagemodels,BRAHMACHARY2025129272,HUANG2024101663,liu2025agenthpo,zhang2023using}. 
Building upon these insights, we present in Appendix~\ref{appendix:LLM_num_opt} empirical evidence showing that (a) LLMs can perform numerical optimization, and (b) that performance is competitive with established methods. This analysis lays the groundwork for our approach. Leveraging this capability, we introduce a simple, task-agnostic prompting strategy for policy search that withholds any semantic or domain-specific information from the LLM. 

\begin{tcolorbox}[colback=tbox_bg,colframe=tbox_frame,title=\props\ Prompt]
{\small
You are a good global optimizer, helping me find the global maximum of a mathematical function f(params).
I will give you the function evaluation and the current iteration number at each step. Your goal is to propose
input values that efficiently lead us to the global maximum within a limited number of iterations (400).

\noindent
\begin{enumerate} [noitemsep,topsep=0pt]
    \item{\makebox[5cm][l]{Regarding the parameters param:} \textcolor{tbox_frame}{\% definitions of parameters}}
    \item{\makebox[5cm][l]{Here's how we'll interact:} \textcolor{tbox_frame}{\% formatting instructions}}
    \item{\makebox[5cm][l]{Remember:} \textcolor{tbox_frame}{\% constraints to be respected}}
\end{enumerate}}
\end{tcolorbox}    

\noindent\begin{minipage}{\textwidth}
\captionof{figure}{Summary of the structure, information, and instructions of a task-agnostic \props\ prompt.}\label{fig:promptProps}
\end{minipage}

Fig.~\ref{fig:promptProps} illustrates a truncated version of the prompt (full prompt in Appendix~\ref{appendix:complete prompts}). The system message specifies the role of the LLM as a global optimizer and indicates the total number of optimization iterations. The prompt includes three key components:
(1) definitions of the parameters to be optimized,
(2) formatting instructions for the LLM’s output, and
(3) any additional constraints the LLM must adhere to during optimization. At each iteration, the LLM receives the prompt $\mathcal{P}$ along with a history $\Gamma$ of previous parameter suggestions and their associated rewards (i.e., in-context examples). It then proposes a new parameter vector $\theta$, accompanied by a textual justification of the update. These justifications add a layer of interpretability to the search process, as they describe observed trends in the data. More specifically, they address how certain parameters or combinations thereof influence the reward, as illustrated in Appendix~\ref{appendix:reasoning example}. Broadly speaking, the justifications can be viewed as \emph{textual gradients}: plain-language explanations of how and why parameters are modified. In this sense, they are conceptually similar to the TextGrad approach introduced in~\citep{yuksekgonul2024textgradautomaticdifferentiationtext}.
 
\subsection{\propss: Semantically-Augmented Prompted Policy Search}
\label{ref:propss}

A core motivation for leveraging LLMs in policy search is their capacity to process and reason over natural language inputs. In \propss, we extend the basic framework to incorporate rich, task-specific, and contextual knowledge into the reinforcement learning process via semantically-informed prompts. The linguistic input can include, for example:
(1) descriptions of the task or environment,
(2) detailed definitions of parameter types,
(3) specifications of the policy structure, and
(4) human-provided hints or constraints regarding optimal behavior.

Fig.~\ref{fig:promptPropsPlus} illustrates the prompt format used for this semantically-augmented variant of \props. The example shown describes the CartPole environment, using text adapted from publicly available documentation (e.g., OpenAI Gym/Gymnasium). In this example, the prompt specifies details such as the task description, action space (binary), policy parameterization (linear), and reward structure. Additionally, it includes optional expert-provided guidance on desirable or undesirable policy behaviors, framed as constraints. To ensure consistency and reduce potential sources of bias, we adopt a standardized prompt structure across all experiments, following the template shown in Fig.~\ref{fig:promptPropsPlus}. While not all semantic information may be useful to the LLM, this uniform presentation enables a fair and interpretable evaluation of how linguistic context influences policy search. The remaining elements of \propss\ are identical to the previously introduced \props\ prompt.

\begin{tcolorbox}[colback=tbox_bg,colframe=tbox_frame,title=\propss\ Prompt]
{\footnotesize
You are a good global RL policy optimizer, helping me find an optimal policy in the following environment:\vspace{0.1cm}

1. {\makebox[5cm][l]{Environment:} \textcolor{tbox_frame}{\% definition of the environment, parameters and policy}}

\textcolor{darkgray}{
In the cartpole environment, a pole is attached by an un-actuated joint to a cart which moves along a frictionless track. The pendulum is placed upright on the cart and the goal is to balance the pole by applying forces in the left and right direction on the cart.The state is a vector of 4 elements, representing the
cart position (-4.8 to 4.8), cart velocity (-inf to inf), pole angle (-0.418 to 0.418 rad), and pole angular velocity (-inf to inf) respectively. The goal is to keep the pole upright and the cart within the bounding position of [-2.4,
2.4]. The action space consists of 2 actions (0: push left, 1: push right).}\vspace{0.15cm}

\textcolor{darkgray}{The policy is a linear policy with 10 parameters and works as follows: action = argmax(...)
The reward is +1 for every time step the pole is upright and the cart is within the bounding position. The
episode ends when the pole falls over or the cart goes out of bounds.}\vspace{0.15cm}

\noindent
\begin{enumerate} [noitemsep,topsep=0pt, leftmargin=*]
    \setcounter{enumi}{1}
    \item{\makebox[5cm][l]{Regarding the parameters param:} \textcolor{tbox_frame}{\% definitions of parameters}}
    \item{\makebox[5cm][l]{Here's how we'll interact:} \textcolor{tbox_frame}{\% formatting instructions}}
    \item{\makebox[5cm][l]{Remember:} \textcolor{tbox_frame}{\% constraints to be respected}}
\end{enumerate}}
\end{tcolorbox}

\noindent\begin{minipage}{\textwidth}
\captionof{figure}{Summary of the structure, information and instructions utilized to construct domain-specific \propss\ prompt, with an example of the prompt for CartPole environment.}\label{fig:promptPropsPlus}
\end{minipage}

\section{Experiments}
We evaluate the performance of both \props\ and \propss, using GPT-4o \citep{openai_chatgpt}, across 15 widely-used reinforcement learning benchmarks from the OpenAI Gym and Gymnasium~\citep{towers2024gymnasium} suites. For tasks with continuous state spaces, we employ linear policy representations, while discrete-state tasks use tabular policies. The selected environments span a diverse range of RL domains, including classic control problems (e.g., CartPole, MountainCar), games (e.g., Pong, Nim), continuous control tasks (e.g., MuJoCo environments \citep{todorov2012mujoco}), etc. Sec.~\ref{sec:exp_basic_props} begins by evaluating the base \props\ method using the numerical optimization prompt, comparing its performance against several state-of-the-art RL baselines. We extend this evaluation to \propss, where prompts are augmented with domain-specific knowledge, and detailed policy specifications, and human provided hints.   Finally, in Sec.~\ref{sec:history} we conclude our evaluation by analyzing four key aspects of our approach:
(a) the effect of context length and in-context history size,
(b) the computational costs of the method,
(c) the choice of underlying LLM, and
(d) the ability to fine-tune a lightweight LLM for \props.

\subsection{Experimental Setup} \label{subsec:experimental setup}

\paragraph{Baselines, Training and Evaluation Metrics.} We compare our approach against seven widely adopted RL algorithms spanning both continuous and discrete action spaces, namely Deep Q-Network (DQN)~\citep{mnih2015human}, Deep Deterministic Policy Gradient (DDPG)~\citep{journals/corr/LillicrapHPHETS15}, Twin Delayed Deep Deterministic Policy Gradient (TD3)~\citep{fujimoto2018addressing}, Soft Actor-Critic (SAC)~\citep{haarnoja2018soft}, Trust Region Policy Optimization (TRPO)~\citep{schulman2015trust}, Proximal Policy Optimization (PPO)~\citep{schulman2017proximal}, and Advantage Actor-Critic (A2C)~\citep{mnih2016asynchronous}.

To ensure a fair comparison across methods, we evaluate all tasks and algorithms over 10 independent trials, with each trial consisting of 8,000 episodes. For our proposed approaches, \props\ and \propss, the LLM is updated every 20 episodes, resulting in 400 optimization iterations per trial. Default LLM hyperparameters are utilized in the policy search. Experiments are carried out with an Intel Xeon W-2125 CPU with 32GB memory. Baseline results are obtained using implementations from the publicly available and community-maintained Stable-Baselines3 and SB3-Contrib libraries~\citep{raffin2021stable}. For tasks and algorithms with pre-tuned hyperparameters provided by these libraries, we adopt those settings directly. In cases where no tuned configuration is available, we use the respective algorithm’s default hyperparameters. Final performance is reported as the mean of the best episodic reward achieved during training, averaged over 10 trials for each method.

\subsection{Evaluation Results}
\begin{wrapfigure}{r}{0.4\textwidth}
\vspace{-12pt}
  \centering
  \includegraphics[trim= 0 0 0 0, clip, width=0.4\textwidth]{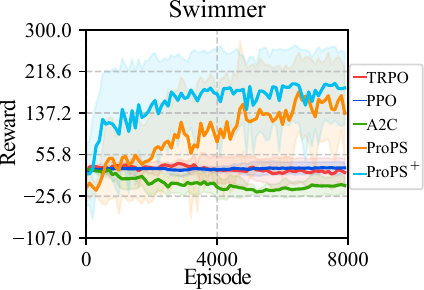}
  \caption{Episodic performance of \props\ and \propss compared to baseline algorithms in the Swimmer task.}
  \label{fig:swimmer_episodic_reward}
  \vspace{-10pt}
\end{wrapfigure}
\textbf{Evaluating \props:}\label{sec:exp_basic_props}
We begin by evaluating the basic variant of our approach, Prompted Policy Search (ProPS), which operates solely on the numerical reward signal without incorporating any semantic information. Results are summarized in Tab.~\ref{Tab:results1}. We noted that across all environments, \props\ consistently improves performance over time. While it underperforms relative to baselines in the \textbf{Walker} and \textbf{Maze} domains, it demonstrates competitive or superior performance in the remaining tasks. Notably, in \textbf{7 out of 15 environments}, \props\ outperforms all baseline algorithms. These results highlight its ability to discover effective policies using only reward-based feedback, despite lacking explicit environment models or gradient-based updates. In the inverted pendulum task, \props\ consistently reaches the global optimum in all experiments. Likewise, in the  \textbf{Swimmer} environment \props\ shows substantial performance gains over other methods. In fact, \props\ significantly outperforms the baselines algorithms throughout the training process, as illustrated in Fig.~\ref{fig:swimmer_episodic_reward}. Note that the results for baseline algorithms are obtained using hyperparameters from the SB3-Contrib libraries~\citep{raffin2021stable}. We provide further analysis our approach in comparison to the baseline algorithms in Appendix~\ref{appendix:continuoustasks}.

\begin{table*}[htb!]
\begin{center}
\setlength{\tabcolsep}{1pt}
\footnotesize
\caption{Performance of \props~in comparison to baseline algorithms in 15 different environments.}
\centering
\scalebox{0.99}{
\label{Tab:results1}
\begin{tabular}{l c c c c c}
    \toprule
    \textbf{Domain} & \textbf{A2C} & \textbf{DQN} & \textbf{PPO} & \textbf{TRPO} & \textbf{\props} (Ours) \\
    \midrule

Mount. Car (C) & 74.10 $\pm$ 7.44 &  N/A & 78.16 $\pm$ 5.32 & 1.18 $\pm$ 3.54 & \cellcolor[HTML]{CCf2FF}\textbf{87.21 $\pm$ 29.28} \\
Inverted Pend. & 155.15 $\pm$ 53.79 &  N/A & 218.65 $\pm$ 129.31 & 571.31 $\pm$ 358.88 & \cellcolor[HTML]{CCf2FF}\textbf{1000.00 $\pm$ 0.00} \\
Inv. Dbl. Pend. & 102.90 $\pm$ 32.04 &  N/A & 108.60 $\pm$ 4.12 & \cellcolor[HTML]{CCf2FF}3609.37 $\pm$ 4000.04 & 128.17 $\pm$ 24.52 \\
Reacher & -53.48 $\pm$ 21.42 &  N/A & \cellcolor[HTML]{CCf2FF}-7.32 $\pm$ 0.38 & -8.93 $\pm$ 1.39 & -11.32 $\pm$ 1.37 \\
Swimmer & 39.40 $\pm$ 6.54 &  N/A & 37.31 $\pm$ 7.19 & 52.96 $\pm$ 18.86 & \cellcolor[HTML]{CCf2FF}\textbf{218.83 $\pm$ 58.45} \\
Hopper & 123.49 $\pm$ 86.46 &  N/A & 351.75 $\pm$ 157.71 & \cellcolor[HTML]{CCf2FF}716.90 $\pm$ 385.20 & 284.16 $\pm$ 165.62 \\
Walker & 355.84 $\pm$ 154.02 &  N/A & 469.78 $\pm$ 159.17 & \cellcolor[HTML]{CCf2FF}519.38 $\pm$ 73.15 & 147.17 $\pm$ 81.20 \\
Frozen Lake & 0.15 $\pm$ 0.03 & 0.13 $\pm$ 0.03 & 0.16 $\pm$ 0.02 & 0.22 $\pm$ 0.05 & \cellcolor[HTML]{CCf2FF}\textbf{0.57 $\pm$ 0.17} \\
Cliff Walking & -172.30 $\pm$ 14.49 & -320.93 $\pm$ 29.60 & -94.35 $\pm$ 3.96 & \cellcolor[HTML]{CCf2FF}-66.60 $\pm$ 13.61 & -100.00 $\pm$ 0.00 \\
Maze & \cellcolor[HTML]{CCf2FF}0.97 $\pm$ 0.00 & 0.86 $\pm$ 0.22 & 0.96 $\pm$ 0.00 & \cellcolor[HTML]{CCf2FF}0.97 $\pm$ 0.00 & 0.55 $\pm$ 0.83 \\
Nim & \cellcolor[HTML]{CCf2FF}0.58 $\pm$ 0.10 & -0.59 $\pm$ 0.03 & 0.39 $\pm$ 0.07 & 0.50 $\pm$ 0.10 & 0.33 $\pm$ 0.29 \\
Mount. Car (D) & -200.00 $\pm$ 0.00 & -194.36 $\pm$ 1.47 & -200.00 $\pm$ 0.00 & -200.00 $\pm$ 0.00 & \cellcolor[HTML]{CCf2FF}\textbf{-126.11 $\pm$ 21.67} \\
Navigation & 3670.87 $\pm$ 124.29 & -127.59 $\pm$ 7.13 & 4127.43 $\pm$ 24.29 & \cellcolor[HTML]{CCf2FF}4223.51 $\pm$ 19.70 & 2587.30 $\pm$ 707.35 \\
Pong & 0.57 $\pm$ 0.14 & 0.56 $\pm$ 0.02 & 2.29 $\pm$ 0.91 & 1.36 $\pm$ 1.05 & \cellcolor[HTML]{CCf2FF}\textbf{2.80 $\pm$ 0.26} \\
Cart Pole & 64.97 $\pm$ 51.68 & 31.22 $\pm$ 1.39 & 365.86 $\pm$ 73.38 & 465.34 $\pm$ 62.32 & \cellcolor[HTML]{CCf2FF}\textbf{478.27 $\pm$ 65.17} \\

\bottomrule
\multicolumn{5}{l}{Means$\pm$standard errors} \\
\end{tabular}}
\end{center}
\vspace{-0.2in}
\end{table*}

\begin{figure}[h!]
    \centering
    \includegraphics[width=\linewidth]{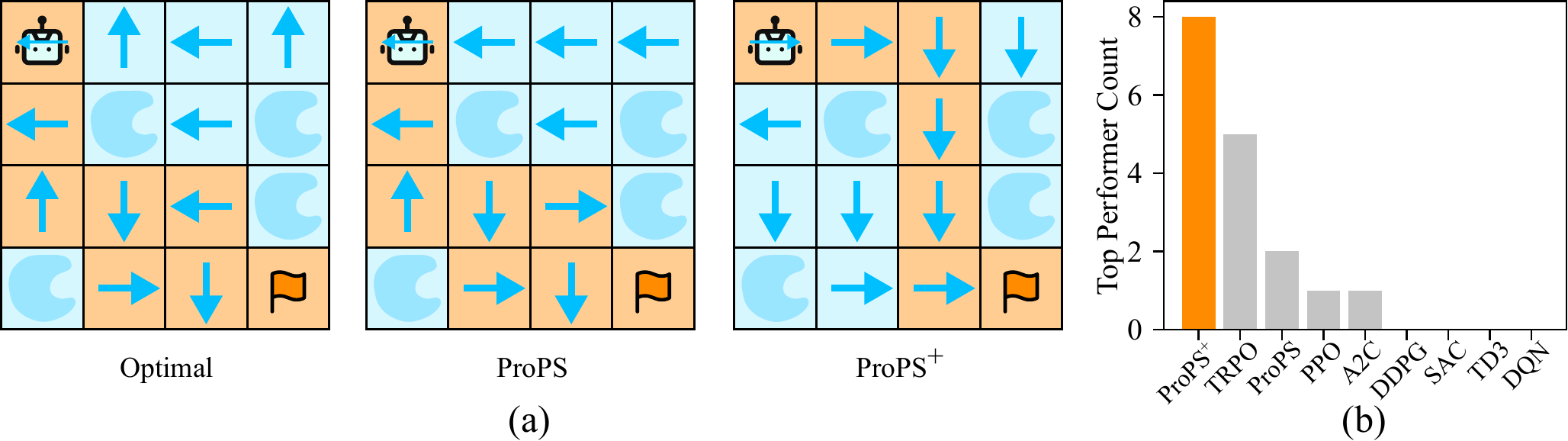}
    \caption{\textbf{(a)} Without semantic information, \props\ is able to learn a successful policy. Note that this policy avoids most of the possibilities of falling into a hole. By contrast, \propss\ is provided with task descriptions, but it created a policy that will only work when the environment is deterministic. \textbf{(b)} \propss\ reaches top performance in 8 out of 15 tasks in our empirical evaluations.}
    \label{fig:frozenlake-top-1-count}
    \vspace{-0.5cm}
\end{figure}

\noindent
\textbf{Performance of \propss}:\label{sec:exp_props_plus} We next evaluate the enhanced variant of our approach, \propss, which incorporates semantic context into the prompt in the form of domain-specific knowledge and explicit policy descriptions. An example prompt is shown in Fig.~\ref{fig:promptPropsPlus}, with content primarily adapted from the official environment documentation provided by Gymnasium~\citep{openai_gym}. Tab.~\ref{Tab:results2} compares the performance of \propss\ to both the base \props\ method and the top two performing RL baselines for each task. For continuous control environments, the top baselines are selected from a pool of six algorithms; for discrete environments, from a set of four. The best-performing method for each task is highlighted in light blue. Overall, we observe that the inclusion of semantic information improves performance in most tasks compared to vanilla \props. Notably, \propss\ outperforms all baselines in several environments, highlighting the value of natural language as a source of auxiliary supervision. However, this trend is not universal. In the \textbf{FrozenLake} environment—characterized by stochastic transitions—the inclusion of task semantics leads to degraded performance. Although the prompt explicitly describes the world as "slippery" (Fig.~\ref{fig:frozenlake-top-1-count}(a)), \propss\ fails to account for this uncertainty and generates a policy under the incorrect assumption of deterministic dynamics. The resulting behavior appears reasonable at first—consistently moving toward the goal—but performs poorly due to the stochasticity of state transitions. In contrast, the base \props\ method, which lacks domain assumptions, performs better by relying solely on observed reward feedback. This example illustrates that while semantic context can enhance performance, it can also introduce \textbf{misleading inductive biases} if domain nuances are misunderstood by the LLM.
\begin{table*}[htb!]
\begin{center}
\footnotesize
\setlength{\tabcolsep}{3pt}
\caption{Comparison of \propss~performance with top baseline algorithms across 15 environments.}
\centering
\scalebox{0.99}{
\label{Tab:results2}
\begin{tabular}{l r l r l l l}
    \toprule
    \textbf{Domain} & \multicolumn{2}{c}{\textbf{Best Baseline}} & \multicolumn{2}{c}{\textbf{2nd Best Baseline}} & \multicolumn{1}{c}{\textbf{\props}} & \multicolumn{1}{c}{\textbf{\propss}}\\
    \midrule

Mount. Car (C) & SAC & 86.65 $\pm$ 0.84 & PPO & 78.16 $\pm$ 5.32 & 87.21 $\pm$ 29.28 & \cellcolor[HTML]{CCf2FF}\textbf{89.16 $\pm$ 29.72} \\
Inverted Pend. & TRPO & 571.31 $\pm$ 358.88 & PPO & 218.65 $\pm$ 129.31 & \cellcolor[HTML]{CCf2FF}\textbf{1000.00 $\pm$ 0.00} & \cellcolor[HTML]{CCf2FF}\textbf{1000.00 $\pm$ 0.00} \\
Inv. Dbl. Pend. & \cellcolor[HTML]{CCf2FF}TRPO & \cellcolor[HTML]{CCf2FF}3609.37 $\pm$ 4000.04 & PPO & 108.60 $\pm$ 4.12 & 128.17 $\pm$ 24.52 & 148.39 $\pm$ 48.65 \\
Reacher & \cellcolor[HTML]{CCf2FF}PPO & \cellcolor[HTML]{CCf2FF}-7.32 $\pm$ 0.38 & TRPO & -8.93 $\pm$ 1.39 & -11.32 $\pm$ 1.37 & -18.15 $\pm$ 22.06 \\
Swimmer & TRPO & 52.96 $\pm$ 18.86 & A2C & 39.40 $\pm$ 6.54 & 218.83 $\pm$ 58.45 & \cellcolor[HTML]{CCf2FF}\textbf{227.30 $\pm$ 56.23} \\
Hopper & \cellcolor[HTML]{CCf2FF}TRPO & \cellcolor[HTML]{CCf2FF}716.90 $\pm$ 385.20 & PPO & 351.75 $\pm$ 157.71 & 284.16 $\pm$ 165.62 & 356.22 $\pm$ 292.35 \\
Walker & \cellcolor[HTML]{CCf2FF}TRPO & \cellcolor[HTML]{CCf2FF}519.38 $\pm$ 73.15 & PPO & 469.78 $\pm$ 159.17 & 147.17 $\pm$ 81.20 & 126.75 $\pm$ 136.44 \\
Frozen Lake & TRPO & 0.22 $\pm$ 0.05 & PPO & 0.16 $\pm$ 0.02 & \cellcolor[HTML]{CCf2FF}\textbf{0.57 $\pm$ 0.17} & 0.19 $\pm$ 0.05 \\
Cliff Walking & \cellcolor[HTML]{CCf2FF}TRPO & \cellcolor[HTML]{CCf2FF}-66.60 $\pm$ 13.61 & PPO & -94.35 $\pm$ 3.96 & -100.00 $\pm$ 0.00 & -96.40 $\pm$ 22.90 \\
Maze & \cellcolor[HTML]{CCf2FF}A2C & \cellcolor[HTML]{CCf2FF}0.97 $\pm$ 0.00 & \cellcolor[HTML]{CCf2FF}TRPO & \cellcolor[HTML]{CCf2FF}0.97 $\pm$ 0.00 & 0.55 $\pm$ 0.83 & \cellcolor[HTML]{CCf2FF}\textbf{0.97 $\pm$ 0.00} \\
Nim & A2C & 0.58 $\pm$ 0.10 & TRPO & 0.50 $\pm$ 0.10 & 0.33 $\pm$ 0.29 & \cellcolor[HTML]{CCf2FF}\textbf{0.97 $\pm$ 0.09} \\
Mount. Car (D) & DQN & -194.36 $\pm$ 1.47 & A2C & -200.00 $\pm$ 0.00 & -126.11 $\pm$ 21.67 & \cellcolor[HTML]{CCf2FF}\textbf{-116.71 $\pm$ 15.20} \\
Navigation & \cellcolor[HTML]{CCf2FF}TRPO & \cellcolor[HTML]{CCf2FF}4223.51 $\pm$ 19.70 & PPO & 4127.43 $\pm$ 24.29 & 2587.30 $\pm$ 707.35 & 2779.55 $\pm$ 270.65 \\
Pong & PPO & 2.29 $\pm$ 0.91 & TRPO & 1.36 $\pm$ 1.05 & 2.80 $\pm$ 0.26 & \cellcolor[HTML]{CCf2FF}\textbf{2.99 $\pm$ 0.03} \\
Cart Pole & TRPO & 465.34 $\pm$ 62.32 & PPO & 365.86 $\pm$ 73.38 & 478.27 $\pm$ 65.17 & \cellcolor[HTML]{CCf2FF}\textbf{500.00 $\pm$ 0.00} \\
     
\bottomrule
\multicolumn{5}{l}{Means$\pm$standard errors} \\
\end{tabular}}
\end{center}
\vspace{-0.2in}
\end{table*}

To quantify overall performance, we count the number of tasks for which each method achieves the highest average episodic reward. As shown in Fig.~\ref{fig:frozenlake-top-1-count}(b), \propss\ achieves \textbf{top performance in 8 out of 15 tasks}, outperforming both TRPO (5 tasks) and \props\ (2 tasks). A striking example is the \textbf{Nim} environment, where vanilla \props\ was initially a low-performing method. With the inclusion of task semantics in \propss, it becomes the top performer—demonstrating the potential of linguistic context to transform search performance when relevant domain knowledge is available.

\begin{figure}[hbt!]
    \centering
    \includegraphics[width=\linewidth]{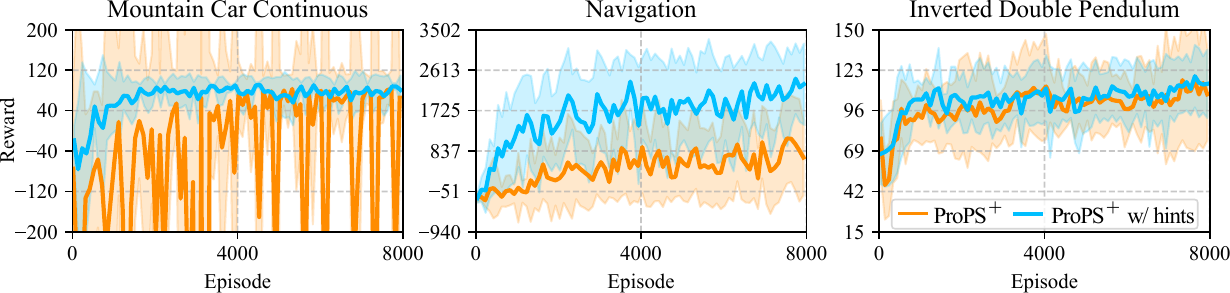}
    \caption{Episodic performance of \propss\ with and without hints across three tasks.}
    \label{fig:propss vs propss with hints}
\end{figure}

\textbf{Performance of \propss\ with Expert Hints:}
\label{sec:exp_props_plus_w_hints}
We also evaluate the impact of human-provided hints about the optimal policy. To this end, we provide hints as part of the prompt in \propss, such as ``\emph{when the velocity is negative, the
force should be negative to push the car back}". Specific hints provided for each of the tasks are described in Appendix \ref{appendix:propsswithhints}. As can be seen in Fig.~\ref{fig:propss vs propss with hints}, we observe that the addition of hints improves our proposed method, leading to faster learning especially at the beginning of the RL process. In both the Mountain Car and Navigation domains, it results in higher early rewards and halves the number of iterations needed for effective policies. In Navigation, hints enable policies to achieve two times the reward of vanilla \props. We hypothesize that hints allow the LLM to initialize policies in a part of the search space that is more amenable to improvement. In tasks such as the Inverted Double Pendulum, it may be harder to express hints and advice in natural language, due to the complex interactions between the cart, the two links and gravity. In such domains, we notice that hints may provide benefits in early exploration but these advantages diminish during the later exploitation phase (as seen in Fig.~\ref{fig:propss vs propss with hints}: Inverted Double Pendulum).  

Table~\ref{Tab:ProPS+ vs ProPS+ w h} reports the changes in the mean of the best episodic rewards during training across six domains. As shown, incorporating informative hints into our approach consistently enhances performance, yielding higher mean rewards and reduced variability across runs. These results further reinforce the positive impact of providing human-guided cues, aligning with our earlier observations on accelerated learning and improved policy search. We also assess the relationship between the quality of a provided hint and its impact on RL performance through an \textbf{ablation study} \textbf{detailed in Appendix~\ref{appendix:ablationstudy}}.

\begin{table*}[htb!]
\begin{center}
\setlength{\tabcolsep}{1pt}
\footnotesize
\caption{Comparison between \propss and \propss with hints in six illustrative domains.}
\centering
\scalebox{0.99}{
\label{Tab:ProPS+ vs ProPS+ w h}
\begin{tabular}{l c c c c c}
    \toprule
    \textbf{Domain} & \textbf{\propss} & \textbf{\propss w hints} & Pct. Change Reward & Pct. Change std\\
    \midrule
Mount. Car (C) & 89.16 $\pm$ 29.72 &  98.70 $\pm$ 0.89 & 10.70\% & -97.01\% \\
Navigation & 2779.55 $\pm$ 270.65 & 3022.66 $\pm$ 135.54 & 8.75\% & -49.92\%\\
Inv. Dbl. Pend. & 148.39 $\pm$ 48.65 &  161.07 $\pm$ 52.28 & 8.54\% & 7.45\% \\
Cliff Walking & -96.4 $\pm$ 22.90 & -54.60 $\pm$ 1.18 & 43.36\% & -94.85\% \\
Reacher & -18.15 $\pm$ 22.06 & -9.66 $\pm$ 1.33 & 46.78\% & -93.97\% \\
Walker & 126.75 $\pm$ 136.44 & 205.47 $\pm$ 91.60 & 62.11\% & -32.86\% \\
\bottomrule
\multicolumn{5}{l}{Means$\pm$standard deviation} \\
\end{tabular}}
\end{center}
\vspace{-0.2in}
\end{table*}

\subsection{Evaluating LLM Models, Runtime, Context Length, and the Effect of Finetuning\label{sec:history}}
\begin{figure}[htb!]
    \centering
    \includegraphics[width=\linewidth]{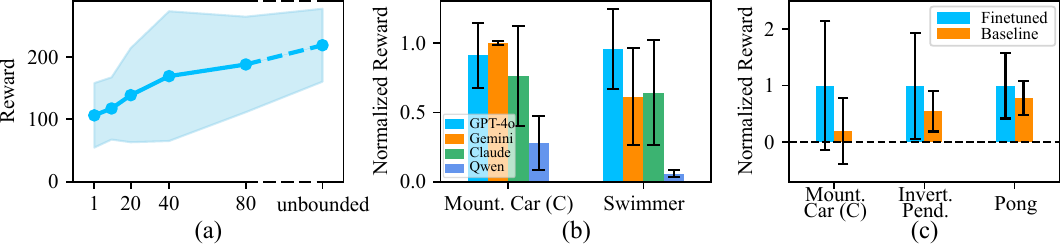}
    \caption{\textbf{(a):} Reward on Mountain Car as a function of history length (x-axis reflects number of in-context examples in history). \textbf{(b):} \props\  performance when using different LLMs. \textbf{(c):} \props\  performance before and after fine-tuning a Qwen2.5-14B-Instruct model.}
    \label{fig:history_buffer_across_llm}
\end{figure}

We conclude our evaluation by investigating the following four aspects of our methodology:

\textbf{Impact of In-Context History Size}
We first examine whether the number of in-context examples (i.e., the history length $N$ iterations) influences policy search performance. Fig.~\ref{fig:history_buffer_across_llm}(a) shows the results on the Mountain Car task. We observe a clear, nearly linear improvement in average reward as $N$ increases. When $N=1$ (which is analogous to a conventional optimizer maintaining only a single candidate) the reward plateaus around 100. In contrast, when the full history is utilized (unbounded N), the agent reaches the maximum reward of 200. This highlights the benefit of leveraging historical parameter-reward pairs, as the LLM is able to synthesize more effective updates over time.

\begin{wrapfigure}{r}{0.38\textwidth}
    \centering
    \includegraphics[width=\linewidth]{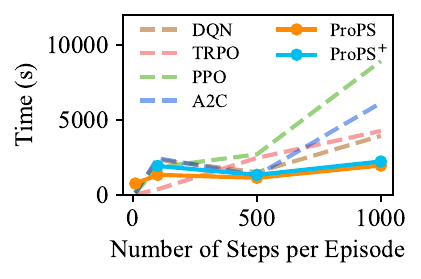}
    \caption{Comparison of runtime: proposed methods (\props\ and \propss) vs. traditional RL baselines as a function of steps per episode.}
    \label{fig:times}
    \vspace{-10pt}
\end{wrapfigure}

\textbf{Run Time Comparison} Next, we evaluate computational efficiency of our proposed methods, \props\ and \propss, in comparison to the baselines. To ensure a fair comparison that accounts for potential differences in CPU utilization during training, we recorded the CPU time for traditional RL algorithms. For \props\ and \propss, the total time measured includes both CPU time and API call duration. Fig.~\ref{fig:times} illustrates these comparisons. We observe that \props\ and \propss\ in this setting show modest time requirements when compared with the baselines. It is important to note that the number of steps per episode often influences baseline computation time differences; the more timesteps required to finish one episode, the longer the computation time needed.

\textbf{Effect of LLM Choice:}
We next assess the robustness of our method across different large language models. Specifically, we evaluate GPT-4o \citep{openai_chatgpt}, Gemini-2.5-Flash \citep{team2023gemini}, Claude-3.7-sonnet~\citep{anthropic_claude_3_7_2025} and Qwen2.5-14B-Instruct~\citep{qwen} on the Mountain Car and Swimmer tasks. As shown in Fig.~\ref{fig:history_buffer_across_llm}(b), all proprietary models show strong performance, demonstrating that modern LLMs are capable of supporting effective prompted policy search, albeit with differences in sample efficiency and final performance. However, by comparison, lightweight LLMs such as Qwen are free and resource-efficient but have more limited capabilities with regards to numerical optimization and policy search. 

\textbf{Fine-Tuning for Policy Search:}
Thus, we explore whether a lightweight LLM can explicitly be fine-tuned to improve its prompted policy search capabilities. This would allow users to eschew proprietary models for smaller, more resource-efficient models to solve challenging problems. To this end, we perform GRPO \citep{shao2024deepseekmath} finetuning of the Qwen2.5-14B-Instruct~\citep{qwen2.5} model\footnote{\url{https://huggingface.co/Qwen/Qwen2.5-14B-Instruct}} using a dataset of $2000$ randomly generated policy parameters with the \props\ prompt for the Mountain Car Continuous task. Here, the reward for GRPO finetuning is the same as the task reward for Mountain Car. After finetuning, we evaluate the fine-tuned model on three tasks: Mountain Car, Inverted Pendulum and Pong to assess generalization. Fig.~\ref{fig:history_buffer_across_llm}(c) shows the fine-tuned model outperforms its pre-trained counterpart on all tasks, suggesting that targeted fine-tuning can enhance general policy search capabilities beyond the training task. We provide the details of fine-tuning in Appendix \ref{appendix:LLM_finetuning}.

\section{Discussion and Conclusion} \label{sec:discussion and limitations}
We introduced Prompted Policy Search, a novel framework that leverages large language models for reinforcement learning. By integrating linguistic reasoning with numerical optimization, \propss\ is able to embed valuable natural language information like prior domain knowledge, goals, or user-provided policy hints directly into the learning process. We demonstrated that LLMs can successfully perform RL on a wide variety of tasks without the need for an external component and that LLMs can be fine-tuned to improve the performance of \propss. We believe that \propss\ paves the way toward a new generation of differentiable, LLM-based RL methods that dynamically evolve with data, model improvements, and real‑time prompting. Below, we outline limitations and considerations that offer opportunities for further research: 

\textbf{What about Deep Reinforcement Learning?} In our experiments, we focused on RL tasks with policy dimensionalities up to 100 parameters, deliberately omitting Deep RL~\cite{mnih2015human} scenarios that involve high-dimensional neural network policies for end-to-end learning from image inputs. This choice was made to establish clear evidence that LLMs can effectively perform policy search for tasks with a moderate number of dimensions. Extending \propss\ to deep RL presents additional challenges, since Deep RL requires discovering optimal representations and may benefit from unsupervised pre-training. Preliminary results with complex policies with one and two hidden layer and several neurons in hidden layers can be found in Appendix~\ref{appendix:scaling_nn}. We have conducted an early investigation of \props\ on neural network policies resulting in promising results, for details see Appendix~\ref{appendix:neural_networks}. However, more research and new methods for enabling LLMs to uncover optimal neural network representations will be needed for Deep RL tasks.

\textbf{On the Origins of Optimization Capabilities.} How do LLMs exhibit zero-shot numerical optimization abilities without explicit training for such tasks? Recent studies and our empirical evidence (see Appendix~\ref{appendix:LLM_num_opt}) indicate that LLMs possess a remarkable capacity for optimization. One possible hypothesis is that during pretraining, LLMs are exposed to vast corpora containing optimization-related content, such as: (a) descriptions of optimization algorithms, (b) RL training logs, (c)
hyperparameter tuning scripts, or (d) optimization tutorials. This exposure may lead to the development of implicit priors about optimization procedures. The phenomenon of emergent abilities has repeatedly been observed in LLMs~\citep{wei2022emergentabilitieslargelanguage}. While these findings are promising, a deeper analysis is required to understand and \emph{explicitly promote} the mechanisms underlying LLM-based optimization.

\noindent
\textbf{Sensitivity of LLM Optimization Capabilities.} In Appendix~\ref{appendix:brittleness_analysis} we investigate the sensitivity of our approach to different prompts. To evaluate the impact we create several variations of the \props\ prompt and the environment descriptions using the Gemini-2.5 Pro model and evaluate them on the Mountain Car Continuous domain. The results show that the LLMs are resilient to the specific phrasing of the prompt. On the other hand, our results indicate that LLMs may be susceptible to the order in which the information is provided. For example, providing long tables at the beginning of the prompt, can hinder their capability to reason about other important information. In fact, the changing the order of how they respond can sometimes have a detrimental impact on the overall results.

\noindent
\textbf{On Policy Representations.} For LLMs to effectively utilize human-provided hints and domain knowledge, they must translate these inputs into precise modifications of policy parameters. This translation hinges on the choice of policy parameterization. Structured and hierarchical representations are particularly advantageous, as they offer clarity and modularity, enabling LLMs to identify and adjust specific components in response to linguistic cues. For future work, we are interested in policy parameterizations that are both expressive and amenable to modification by LLMs.

\noindent
\textbf{Broader Impacts.} \props\ and \propss\ enable natural language communication between users and optimizers, placing the user at the heart of the optimization. This provides transparency in optimization and facilitates the integration of prior knowledge into the optimizer: an essential feature for domains requiring transparency, safety, and human oversight, e.g., healthcare, robotics, space applications, or rescue operations.

\begin{ack}
This research was supported by a grant from Procter \& Gamble.
\end{ack}

\bibliographystyle{plainnat}
\bibliography{references}

\clearpage
\appendix
\begin{center}
    \LARGE\bfseries Appendix 
\end{center}
\startcontents[apx]

\section*{Table of Contents} 
\printcontents[apx]{}{0}{} 
\clearpage 


\section{LLMs Performing Numerical Optimization}\label{appendix:LLM_num_opt}

Reinforcement Learning (RL) often involves policy search, a process that can be effectively framed as a numerical optimization problem where policy parameters are adjusted to maximize an objective like episodic reward. In our main work, we explore the application of Large Language Models (LLMs) to tackle these complex policy search tasks. To rigorously ground this exploration and better understand the fundamental optimization capabilities of LLMs, this section focuses on evaluating their performance on a suite of standard, ``pure'' numerical optimization benchmarks.

The primary objective here is to systematically assess how well LLMs, specifically Gemini-1.5-pro and GPT-4o, can perform as direct numerical optimizers when tasked with minimizing mathematical functions. We aim to quantify their solution quality in comparison to established classical optimization algorithms (Gradient Descent, Adam, Nelder-Mead) and a Random Search baseline. The insights gained will provide a foundational understanding of their strengths, weaknesses, and potential as general-purpose numerical optimizers, which is directly relevant to their broader application, including the policy optimization challenges addressed in this paper.

\subsection{LLM-based Optimization Approaches}
Our LLM-based optimization strategy employs an iterative process where the LLM actively proposes candidate solutions. This approach is consistent with the general framework used for policy search in the main body of this paper. For every iteration $t$ of the optimization process:

\begin{itemize}
    \item Input to LLM: The LLM (either Gemini-1.5-pro or GPT-4o) is provided with the last evaluated points and their corresponding objective function values $(\boldsymbol{x}_{t-1}, f(\boldsymbol{x}_{t-1}))$. It also receives the current iteration number $t$.
    \item LLM Task: The LLM is prompted to generate a new D-dimensional input vector $\boldsymbol{x}_t$ (where D is the dimensionality of the current objective function).
    \item External Evaluation: The proposed vector $\boldsymbol{x}_t$ is evaluated using the true objective function to obtain its value $f(\boldsymbol{x}_t)$.
    \item Feedback: The new pair $(\boldsymbol{x}_{t}, f(\boldsymbol{x}_{t}))$ is fed back to the LLM for the next iteration.
\end{itemize}
Adam optimizer is employed for the initial two optimization steps as a warmup. Subsequently, the LLM-based optimization is repeated for an additional 98 steps, resulting in a total of 100 optimization steps. The system prompt guiding this interaction is provided below:

\begin{tcolorbox}[colback=tbox_bg,colframe=tbox_frame,title=LLM Numerical Optimization Prompt]
You are an optimization assistant, helping me find the global minimum of a mathematical function.  I will give you the function evaluation and the current iteration number at each step. Your goal is to propose input values that efficiently lead us to the global minimum within a limited number of iterations (100).

\noindent
\begin{enumerate} [noitemsep,topsep=0pt]
    \item{\makebox[5cm][l]{Here's how we'll interact:} \textcolor{tbox_frame}{\% formatting instructions}}
    \item{\makebox[5cm][l]{Remember:} \textcolor{tbox_frame}{\% constraints to be respected}}
\end{enumerate}
\end{tcolorbox}    

\noindent\begin{minipage}{\textwidth}
\captionof{figure}{Summary of the structure, information, and instructions utilized to construct the numerical optimization prompt.}\label{fig:prompt_numoptim}
\end{minipage}

\subsection{Experiment Setup}
\subsubsection{Objective Functions}

To evaluate the optimization capabilities of LLMs and baselines, we selected five standard benchmark objective functions commonly used in the optimization literature. For each function f(x) where x is a D-dimensional vector, the goal is to find $x^* = \text{argmin} f(x)$.

\begin{itemize}
    \item \textbf{Ackley.} $f(x) = -a \exp\left(-b \sqrt{\frac{1}{D} \sum_{i=1}^{D} x_i^2}\right) - \exp\left(\frac{1}{D} \sum_{i=1}^{D} \cos(c x_i)\right) + a + \exp(1)$, where $a = 20, b = 0.2, c = 2\pi$. Its landscape features a large, nearly flat outer region and a central hole where the global minimum lies.
x    \item \textbf{Rastrigin.} $f(x) = 10D + \sum_{i=1}^{D} [x_i^2 - 10 \cos(2\pi x_i)]$. A classic function with a regular, grid-like arrangement of local minima. The global minimum is at f(0, ..., 0) = 0. It is challenging due to the vast number of local optima.

    \item \textbf{Levy.} $f(x) = \sin^2(\pi w_1) + \sum_{i=1}^{D-1} (w_i-1)^2 [1 + 10\sin^2(\pi w_i + 1)] + (w_D-1)^2 [1 + \sin^2(2\pi w_D)]$, where $w_i = 1 + (x_i - 1)/4$.  It has many local minima. The global minimum is at $(x_0, x_1, ...)$ where $x_i = 1$ for all $i$.
    \item \textbf{Weierstrass.} $f(x) = \sum_{i=1}^{D} \left( \sum_{k=0}^{k_{max}} a^k \cos(2\pi b^k (x_i + 0.5)) \right) - D \sum_{k=0}^{k_{max}} a^k \cos(\pi b^k)$, where $a = 0.5, b = 3, k_{max} = 20$. It is continuous everywhere but differentiable nowhere. It presents a highly rugged and fractal-like landscape.
    \item \textbf{Salomon.} $f(x) = 1 - \cos\left(2\pi \sqrt{\sum_{i=1}^{D} x_i^2}\right) + 0.1 \sqrt{\sum_{i=1}^{D} x_i^2}$. It has a ``funnel'' shape with concentric rings of local minima.
\end{itemize}
For each of these five objective functions, experiments were conducted in $D =$ $2$, $4$, $8$, and $16$ dimensions. This variation in dimensionality allows us to assess the scalability and robustness of the different optimization approaches across problems of increasing complexity, resulting in a total of 20 unique optimization tasks.

To prevent optimizers from trivially exploiting easy optima locations (often at the origin), we introduced a shift. For each task (function and dimensionality combination), a random offset vector $\boldsymbol{o}$ was generated, where each component $o_i$ was sampled uniformly from $(0, 20)$. The evaluated function is then $f'(\boldsymbol{x}) = f(\boldsymbol{x} - \boldsymbol{o})$. 

\subsubsection{Baseline Optimizers}
To benchmark the performance of the LLM-based approaches, we compared them against several widely recognized classical optimization algorithms and a simple random search baseline. These baselines were chosen to represent a spectrum of optimization strategies, from gradient-based methods to direct search and uninformed exploration, which are listed below:

\paragraph{Gradient Descent (GD)~\citep{curry1944method}.} A fundamental first-order iterative optimization algorithm. At each step, it moves in the direction opposite to the gradient of the objective function. A manual simple grid search is performed to identify the learning rate of 0.005 in our experiments.
\paragraph{Adam~\citep{adam}.} An adaptive learning rate optimization algorithm that computes individual adaptive learning rates for different parameters from estimates of first and second moments of the gradients. Similar to GD, a simple grid search is conducted to identify the learning rate of 0.5 in our experiments.
\paragraph{Nelder-Mead (NM)~\citep{nelder1965simplex}.} A direct search method that does not require gradient information. It maintains a simplex of D+1 points in D dimensions and iteratively modifies it using operations like reflection, expansion, contraction, and shrinking.
\paragraph{Random Search.} A local search method where, at each iteration, a random perturbation (delta) randomly sampled from a normal distribution $\mathcal{N}(0, 0.3)$ is added to the current best point to generate a new candidate solution. 

\subsubsection{Experiment Details}
Every optimizer is evaluated 50 times for each objective function. We randomly generate 50 initial function input points as the optimization initialization, and share them across all optimizers, which makes sure there is a fair comparison. Every optimization process is conducted for 100 time steps, and the 100th step's result is extracted as the optimization result, as shown below:

\subsection{Results}

\begin{wrapfigure}{r}{0.33\textwidth}
    \centering
    \includegraphics[width=\linewidth]{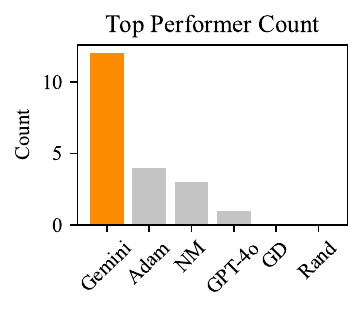}
    \caption{In the 20 numerical optimization tasks, Gemini is able to achieve top-1 performance in 12 tasks.}
    \label{fig:num_optim_top_1}
\end{wrapfigure}

The comprehensive results for all optimizers, presented as the mean and standard deviation of final objective function values from 50 independent runs (each with 100 optimization steps), are detailed in Tab.~\ref{tab:optimization_brief}. An overarching summary, visualized in Fig.~\ref{fig:num_optim_top_1}, reveals that LLM-based optimizers demonstrated notable efficacy. Specifically, Gemini-1.5-pro achieved the best (lowest mean) objective value in a majority of cases, succeeding in 12 out of the 20 distinct optimization tasks. This strong performance was 
followed by Adam, which obtained top results in 4 tasks, Nelder-Mead in 3 tasks, and GPT-4o in 1 task. A closer examination of performance across the specific objective functions highlights varying strengths. Gemini-1.5-pro consistently outperformed all baseline methods across all tested dimensions (2D, 4D, 8D, and 16D) for both the Ackley and Salomon functions. Similarly in Rastrigin function, it achieved the top performance in 2D, 4D, and 8D; in the more challenging 16D Rastrigin task, Adam obtained the best result ($512.17 \pm 145.34$), with Gemini-1.5-pro yielding a competitive second-best performance ($566.40 \pm 139.35$). In contrast, for Weierstrass function, Nelder-Mead algorithm proved superior in the 4D, 8D, and 16D settings, though either Gemini-1.5-pro or GPT-4o ranks as the next best-performing algorithms in these scenarios, outperforming gradient-based methods. On the Levy function, while both Gemini-1.5-pro and GPT-4o achieved significant optimization well beyond the Random Search baseline, Adam emerged as the top performer in the 4D, 8D, and 16D tasks. Gemini-1.5-pro, however, does obtain the leading result in the 2D Levy experiments. Collectively, these findings suggest that LLMs, particularly Gemini-1.5-pro, can be highly competitive, and in many cases superior, to classical optimization techniques on a range of numerical benchmarks. The complete results are shown in Tab.~\ref{tab:optimization}.



\begin{table*}[htb!]
\begin{center}
\footnotesize
\setlength{\tabcolsep}{3pt}
\caption{Comparison of baselines against the Gemini-1.5-Pro and GPT-4o results. Blue cell highlights the best result for the function and the dimension.}
\centering
\scalebox{0.99}{
\label{tab:optimization_brief}
\begin{tabular}{l l r l r l l l}
    \toprule
    \textbf{Function} & \textbf{Dim} & \multicolumn{2}{c}{\textbf{Best Baseline}} & \multicolumn{2}{c}{\textbf{2nd Best Baseline}} & \multicolumn{1}{c}{\textbf{Gemini}} & \multicolumn{1}{c}{\textbf{GPT-4o}}\\
    \midrule
\multirow{4}{*}{Ackley} & 2D & Adam & 14.08 $\pm$ 6.21 & NM & 14.12 $\pm$ 6.45 & \cellcolor[HTML]{CCf2FF}\textbf{10.67 $\pm$ 6.11} & 12.62 $\pm$ 7.15 \\
                       & 4D & Adam & 14.91 $\pm$ 4.80 & NM & 15.62 $\pm$ 3.32 & \cellcolor[HTML]{CCf2FF}\textbf{12.75 $\pm$ 4.82} & 15.24 $\pm$ 4.39 \\
                       & 8D & Adam & 15.27 $\pm$ 2.75 & NM & 15.75 $\pm$ 2.61 & \cellcolor[HTML]{CCf2FF}\textbf{12.41 $\pm$ 3.31} & 16.50 $\pm$ 1.82 \\
                       & 16D & Adam & 15.79 $\pm$ 1.17 & NM & 16.56 $\pm$ 0.99 & \cellcolor[HTML]{CCf2FF}\textbf{13.83 $\pm$ 2.19} & 17.00 $\pm$ 1.09 \\ \midrule
\multirow{4}{*}{Rastrigin} & 2D & NM & 47.68 $\pm$ 40.11 & Adam & 73.77 $\pm$ 59.93 & \cellcolor[HTML]{CCf2FF}\textbf{17.36 $\pm$ 21.83} & 18.78 $\pm$ 31.29 \\
                           & 4D & Adam & 139.13 $\pm$ 74.96 & NM & 204.68 $\pm$ 109.28 & \cellcolor[HTML]{CCf2FF}\textbf{98.88 $\pm$ 71.78} & 125.63 $\pm$ 102.38 \\
                           & 8D & Adam & 263.95 $\pm$ 107.02 & NM & 421.61 $\pm$ 193.26 & \cellcolor[HTML]{CCf2FF}\textbf{204.98 $\pm$ 96.72} & 323.04 $\pm$ 134.70 \\
                           & 16D & \cellcolor[HTML]{CCf2FF}\textbf{Adam} & \cellcolor[HTML]{CCf2FF}\textbf{512.17 $\pm$ 145.34} & NM & 925.40 $\pm$ 257.00 & 566.40 $\pm$ 139.35 & 831.16 $\pm$ 215.72 \\ \midrule
\multirow{4}{*}{Levy}  & 2D & Adam & 13.93 $\pm$ 8.70 & NM & 14.21 $\pm$ 8.81 & \cellcolor[HTML]{CCf2FF}\textbf{5.36 $\pm$ 10.12} & 13.35 $\pm$ 7.78 \\
                       & 4D & \cellcolor[HTML]{CCf2FF}\textbf{Adam} & \cellcolor[HTML]{CCf2FF}\textbf{22.71 $\pm$ 10.07} & GD & 23.14 $\pm$ 9.93 & 23.81 $\pm$ 29.38 & 25.36 $\pm$ 13.65 \\
                       & 8D & \cellcolor[HTML]{CCf2FF}\textbf{Adam} & \cellcolor[HTML]{CCf2FF}\textbf{37.43 $\pm$ 14.66} & GD & 38.20 $\pm$ 14.43 & 68.39 $\pm$ 37.11 & 53.52 $\pm$ 22.85 \\
                       & 16D & \cellcolor[HTML]{CCf2FF}\textbf{Adam} & \cellcolor[HTML]{CCf2FF}\textbf{73.99 $\pm$ 22.80} & GD & 75.40 $\pm$ 22.36 & 191.26 $\pm$ 68.28 & 203.30 $\pm$ 101.14 \\ \midrule
\multirow{4}{*}{Weierstrass}   & 2D & NM & 0.62 $\pm$ 0.78 & Adam & 4.05 $\pm$ 1.13 & 1.88 $\pm$ 0.98 & \cellcolor[HTML]{CCf2FF}\textbf{0.57 $\pm$ 0.86} \\
                               & 4D & \cellcolor[HTML]{CCf2FF}\textbf{NM} & \cellcolor[HTML]{CCf2FF}\textbf{2.62 $\pm$ 1.34} & Adam & 8.06 $\pm$ 1.71 & 4.37 $\pm$ 1.73 & 3.67 $\pm$ 1.81 \\
                               & 8D & \cellcolor[HTML]{CCf2FF}\textbf{NM} & \cellcolor[HTML]{CCf2FF}\textbf{6.44 $\pm$ 2.04} & Adam & 15.99 $\pm$ 2.56 & 10.34 $\pm$ 2.93 & 11.42 $\pm$ 2.24 \\
                               & 16D & \cellcolor[HTML]{CCf2FF}\textbf{NM} & \cellcolor[HTML]{CCf2FF}\textbf{19.77 $\pm$ 4.67} & GD & 31.07 $\pm$ 2.43 & 25.53 $\pm$ 4.36 & 28.86 $\pm$ 3.64 \\ \midrule
\multirow{4}{*}{Salomon}   & 2D & GD & 0.87 $\pm$ 0.40 & NM & 0.87 $\pm$ 0.41 & \cellcolor[HTML]{CCf2FF}\textbf{0.69 $\pm$ 0.31} & 0.81 $\pm$ 0.53 \\
                           & 4D & Adam & 1.27 $\pm$ 0.64 & GD & 1.36 $\pm$ 0.52 & \cellcolor[HTML]{CCf2FF}\textbf{0.82 $\pm$ 0.26} & 1.45 $\pm$ 0.57 \\
                           & 8D & GD & 2.12 $\pm$ 0.49 & NM & 2.14 $\pm$ 0.49 & \cellcolor[HTML]{CCf2FF}\textbf{1.59 $\pm$ 0.38} & 2.19 $\pm$ 0.48 \\
                           & 16D & GD & 2.96 $\pm$ 0.43 & NM & 2.98 $\pm$ 0.43 & \cellcolor[HTML]{CCf2FF}\textbf{2.20 $\pm$ 0.50} & 2.94 $\pm$ 0.49 \\

\bottomrule
\multicolumn{5}{l}{Means$\pm$standard errors} \\
\end{tabular}}
\end{center}
\vspace{-0.1in}
\end{table*}

\clearpage

\begin{landscape}
\captionof{table}{Detailed results for four different baselines in comparison to \props\ based numerical optimization using Gemini-1.5-Pro and GPT-4o. Blue cell represents the best result achieved for a specific function and the dimension.}
\begin{center}
\setlength{\tabcolsep}{1pt}
\footnotesize
\centering
\scalebox{0.99}{
\label{tab:optimization}
\begin{tabular}{l l c c c c c c c}
    \toprule
    \textbf{Function $f$} & \textbf{Dim} & \textbf{Initialization} & \textbf{GD} & \textbf{Adam} & \textbf{Nelder Mead (NM)} & \textbf{Random (Rand)} & \textbf{Gemini} & \textbf{GPT-4o} \\
    \midrule
\multirow{4}{*}{Ackley} & 2D & 17.03 $\pm$ 4.08 & 15.35 $\pm$ 4.27 & 14.08 $\pm$ 6.21 & 14.12 $\pm$ 6.45 & 17.02 $\pm$ 4.13 & \cellcolor[HTML]{CCf2FF}\textbf{10.67 $\pm$ 6.11} & 12.62 $\pm$ 7.15 \\
                       & 4D & 17.60 $\pm$ 2.81 & 15.87 $\pm$ 2.88 & 14.91 $\pm$ 4.80 & 15.62 $\pm$ 3.32 & 17.54 $\pm$ 2.79 & \cellcolor[HTML]{CCf2FF}\textbf{12.75 $\pm$ 4.82} & 15.24 $\pm$ 4.39 \\
                       & 8D & 17.48 $\pm$ 1.88 & 15.88 $\pm$ 1.84 & 15.27 $\pm$ 2.75 & 15.75 $\pm$ 2.61 & 17.48 $\pm$ 1.81 & \cellcolor[HTML]{CCf2FF}\textbf{12.41 $\pm$ 3.31} & 16.50 $\pm$ 1.82 \\
                       & 16D & 17.73 $\pm$ 0.93 & 16.72 $\pm$ 0.97 & 15.79 $\pm$ 1.17 & 16.56 $\pm$ 0.99 & 17.75 $\pm$ 0.91 & \cellcolor[HTML]{CCf2FF}\textbf{13.83 $\pm$ 2.19} & 17.00 $\pm$ 1.09 \\ \midrule
\multirow{4}{*}{Rastrigin} & 2D & 111.99 $\pm$ 72.65 & 91.16 $\pm$ 73.36 & 73.77 $\pm$ 59.93 & 47.68 $\pm$ 40.11 & 108.63 $\pm$ 72.42 & \cellcolor[HTML]{CCf2FF}\textbf{17.36 $\pm$ 21.83} & 18.78 $\pm$ 31.29 \\
                           & 4D & 302.03 $\pm$ 135.81 & 257.86 $\pm$ 133.96 & 139.13 $\pm$ 74.96 & 204.68 $\pm$ 109.28 & 300.86 $\pm$ 142.52 & \cellcolor[HTML]{CCf2FF}\textbf{98.88 $\pm$ 71.78} & 125.63 $\pm$ 102.38 \\
                           & 8D & 618.41 $\pm$ 206.82 & 533.25 $\pm$ 201.18 & 263.95 $\pm$ 107.02 & 421.61 $\pm$ 193.26 & 617.75 $\pm$ 206.92 & \cellcolor[HTML]{CCf2FF}\textbf{204.98 $\pm$ 96.72} & 323.04 $\pm$ 134.70 \\
                           & 16D & 1131.86 $\pm$ 284.23 & 960.97 $\pm$ 274.19 & \cellcolor[HTML]{CCf2FF}\textbf{512.17 $\pm$ 145.34} & 925.40 $\pm$ 257.00 & 1134.37 $\pm$ 293.09 & 566.40 $\pm$ 139.35 & 831.16 $\pm$ 215.72 \\ \midrule
\multirow{4}{*}{Levy}  & 2D & 46.90 $\pm$ 43.54 & 14.28 $\pm$ 8.58 & 13.93 $\pm$ 8.70 & 14.21 $\pm$ 8.81 & 46.54 $\pm$ 44.86 & \cellcolor[HTML]{CCf2FF}\textbf{5.36 $\pm$ 10.12} & 13.35 $\pm$ 7.78 \\
                       & 4D & 103.76 $\pm$ 72.30 & 23.14 $\pm$ 9.93 & \cellcolor[HTML]{CCf2FF}\textbf{22.72 $\pm$ 10.07} & 23.45 $\pm$ 10.84 & 105.51 $\pm$ 75.68 & 23.81 $\pm$ 29.38 & 25.36 $\pm$ 13.65 \\
                       & 8D & 202.06 $\pm$ 87.14 & 38.20 $\pm$ 14.43 & \cellcolor[HTML]{CCf2FF}\textbf{37.43 $\pm$ 14.66} & 54.30 $\pm$ 51.66 & 203.21 $\pm$ 88.09 & 68.39 $\pm$ 37.11 & 53.52 $\pm$ 22.85 \\
                       & 16D & 434.97 $\pm$ 146.39 & 75.40 $\pm$ 22.36 & \cellcolor[HTML]{CCf2FF}\textbf{73.99 $\pm$ 22.80} & 167.79 $\pm$ 86.29 & 443.48 $\pm$ 163.90 & 191.26 $\pm$ 68.28 & 203.30 $\pm$ 101.14 \\ \midrule
\multirow{4}{*}{Weierstrass}   & 2D & 4.29 $\pm$ 1.22 & 4.25 $\pm$ 0.99 & 4.05 $\pm$ 1.13 & 0.62 $\pm$ 0.78 & 4.10 $\pm$ 1.30 & 1.88 $\pm$ 0.98 & \cellcolor[HTML]{CCf2FF}\textbf{0.57 $\pm$ 0.86} \\
                               & 4D & 8.34 $\pm$ 1.65 & 8.12 $\pm$ 1.36 & 8.06 $\pm$ 1.71 & \cellcolor[HTML]{CCf2FF}\textbf{2.62 $\pm$ 1.34} & 8.18 $\pm$ 1.68 & 4.37 $\pm$ 1.73 & 3.67 $\pm$ 1.81 \\
                               & 8D & 15.99 $\pm$ 2.00 & 16.04 $\pm$ 1.92 & 15.99 $\pm$ 2.56 & \cellcolor[HTML]{CCf2FF}\textbf{6.44 $\pm$ 2.04} & 16.36 $\pm$ 2.16 & 10.34 $\pm$ 2.93 & 11.42 $\pm$ 2.24 \\
                               & 16D & 31.17 $\pm$ 3.39 & 31.07 $\pm$ 2.43 & 31.36 $\pm$ 3.77 & \cellcolor[HTML]{CCf2FF}\textbf{19.77 $\pm$ 4.67} & 31.97 $\pm$ 2.76 & 25.53 $\pm$ 4.36 & 28.86 $\pm$ 3.64 \\ \midrule
\multirow{4}{*}{Salomon}   & 2D & 1.81 $\pm$ 0.83 & 0.87 $\pm$ 0.40 & 0.88 $\pm$ 0.42 & 0.87 $\pm$ 0.41 & 1.97 $\pm$ 0.78 & \cellcolor[HTML]{CCf2FF}\textbf{0.69 $\pm$ 0.31} & 0.81 $\pm$ 0.53 \\
                           & 4D & 2.27 $\pm$ 0.88 & 1.36 $\pm$ 0.52 & 1.27 $\pm$ 0.64 & 1.38 $\pm$ 0.53 & 2.44 $\pm$ 0.75 & \cellcolor[HTML]{CCf2FF}\textbf{0.82 $\pm$ 0.26} & 1.45 $\pm$ 0.57 \\
                           & 8D & 3.21 $\pm$ 0.87 & 2.12 $\pm$ 0.49 & 3.77 $\pm$ 4.28 & 2.14 $\pm$ 0.49 & 3.26 $\pm$ 0.80 & \cellcolor[HTML]{CCf2FF}\textbf{1.59 $\pm$ 0.38} & 2.19 $\pm$ 0.48 \\
                           & 16D & 4.04 $\pm$ 0.81 & 2.96 $\pm$ 0.43 & 3.40 $\pm$ 2.20 & 2.98 $\pm$ 0.43 & 4.15 $\pm$ 0.93 & \cellcolor[HTML]{CCf2FF}\textbf{2.20 $\pm$ 0.50} & 2.94 $\pm$ 0.49 \\

\bottomrule
\multicolumn{5}{l}{Means$\pm$standard errors} \\
\end{tabular}}
\end{center}
\end{landscape}
\clearpage

\section{Finetuning LLMs for Reinforcement Learning}\label{appendix:LLM_finetuning}
From the open lightweight models \footnote{\url{https://github.com/eugeneyan/open-llms}}, we chose Qwen2.5 model series \citep{qwen2.5}, as they were finetuned to improve general purpose capabilities. We specifically chose the instruct version as even the smaller models (such as the 3B and 7B) were able to follow instructions and return policy parameters of correct size and rank for different domains. Although they were not able to reason about the parameters even after finetuning. Preliminary results for Qwen2.5 14B model are presented in the paper and the details of finetuning with some extra results are presented here.

The Qwen2.5 14B model was finetuned with a dataset of 2000 data-points with a \props\ prompt, with an example of policy parameters and the rewards shown below. The policy parameters are randomly generated between $[-6.0, 6.0]$ and evaluated for $20$ Mountain car continuous episodes. The finetuning was performed for 5 epochs.

\begin{tcolorbox}[enhanced, breakable, colback=tbox_bg,colframe=tbox_frame,title=\props\ Prompt for Finetuning]
You are good global optimizer, helping me find the global maximum of a mathematical function f(params).\\
I will give you the function evaluation and the current iteration number at each step.\\
Your goal is to propose input values that efficiently lead us to the global maximum within a limited number of iterations (100).\\
\\
\# Regarding the parameters **params**:\\
**params** is an array of 3 float numbers.\\
**params** values are in the range of [-6.0, 6.0] with 1 decimal place.\\
\\
\# Here's how we'll interact:\\
1. I will first provide MAX\_STEPS (100) along with a few training examples.\\
2. You will provide your response in the following exact format:\\
    * Line 1: a new input 'params[0]: ; params[1]: ; params[2]: ', aiming to maximize the function's value f(params).\\
    Please propose params values in the range of [-6.0, 6.0], with 1 decimal place.
    * Line 2: detailed explanation of why you chose that input.\\
3. I will then provide the function's value f(params) at that point, and the current iteration.\\
4. We will repeat steps 2-3 until we reach the maximum number of iterations.\\

\# Remember:\\
1. **Do not propose previously seen params.**\\
2. **The global optimum should be around 100.** If you are below that, this is just a local optimum. You should explore instead of exploiting.\\
3. Search both positive and negative values. **During exploration, use search step size of 1.0**.\\
\\
Next, you will see examples of params and f(params) pairs.\\
params[0]: 3.24; params[1]: 1.72; params[2]: 2.69; f(params): -366.92\\
\\
Now you are at iteration 1 out of 100. Please provide the results in the indicated format. Do not provide any additional texts.
\end{tcolorbox}

Fine-tuning was performed using GRPO \citep{shao2024deepseekmath}. We used GRPO, instead of SFT, because of two reasons:
\begin{enumerate}
        \item Parameter generation problem can be considered a reasoning problem in the policy space, thus the initial success of DeepSeek for mathematics reasoning we chose the GRPO method for finetuning.
        \item During initial examination of the proprietary models, it was found that parameter search would improve if the models were asked to perform two separate steps, present the parameters and think about the reasons for the parameters as well. GRPO allows for a similar structure, as it divides the response generation into two steps -- think and final answer.
\end{enumerate}

The following prompt was used for finetuning with GRPO:
\begin{tcolorbox}[enhanced, breakable, colback=tbox_bg,colframe=tbox_frame,title=GRPO Fine-Tuning Prompt]
A conversation between User and Assistant. The User is looking for a linear control policy for the continuous Mountain Car Domain. Assistant first thinks about the reasoning process in the mind and then provides a policy to the user. The reasoning process and the policy are enclosed within the <think> </think> and <policy> </policy> tags respectively, i.e. <think> reasoning process here </think><policy> policy here </policy>.
\end{tcolorbox}

Four reward functions were used for improving fine-tuning direction:
\begin{enumerate}
    \item Format Reward -- which checks whether the structure of the response follows <think> </think> <policy> </policy> structure or not.
    \item Strict format reward -- which checks format as <think> </think> <policy> </policy> and also checks whether parameters exist between policy tag and the size of the params matches the rank of the domain (mountain car continuous, 3), i.e. <policy>params [0]; params[1]  ; params[2] ;</policy>.
    \item Policy reward -- Parses the response from the model, retrieves the policy parameters, and evaluates the policy. For any reward less than $-200$, is converted to $0$, and any reward greater than $100$ is converted $1$. Reward value between $[-200, 100]$ is normalized between $[0, 1]$.
    \item Policy gradient reward -- The new reward value calculated from policy reward with value between $[0, 1]$ ($r_{new}$), is compared against the normalized reward given in the prompt ($r_{ini}$) and gradient is calculated as -- $Reward = (r_{new} - r_{ini})*r_{new}/r_{ini}$. 
\end{enumerate}

An A100 GPU was used for the inference cycles, and fine-tuning was performed with 2 H100 GPUs. The finetuning was performed using the LoRA framework \citep{hu2022lora}, and we chose $5$ layers only spread across the attention, and the linear layers. Other hyper-parameters for finetuning --
\begin{itemize}
    \item Learning rate -- $1e^{-5}$.
    \item 4 parallel generations for GRPO.
    \item Max completion length -- $256$.
    \item Gradient accumulation steps -- $16$.
\end{itemize}
The Transformer Reinforcement Learning (TRL) version of GRPO implementation\footnote{\url{https://github.com/huggingface/trl/blob/main/trl/trainer/grpo_trainer.py}} was used for finetuning. The complete code is provided with the code part of the submission.

\begin{table*}[htb!]
\begin{center}
\footnotesize
\setlength{\tabcolsep}{3pt}
\caption{Comparison of Baseline and Finetune models for Qwne2.5 14B for untrained domains, i.e. Pong and Inverted pendulum.}
\centering
\scalebox{0.99}{
\label{tab:finetune_results}
\begin{tabular}{l l c}
\toprule
\textbf{Domain} & \textbf{Model}  & \textbf{Average Max Rewards} \\ \midrule
\multirow{2}{*}{Mountain Car (Cont.)} & Baseline & 8.11 $\pm$ 24.45 \\
    & Finetuned & 42.01 $\pm$ 48.1 \\ \midrule
\multirow{2}{*}{Inverted Pendulum} & Baseline & 55.64 $\pm$ 36.5 \\
    & Finetuned & 101.148 $\pm$ 94.804 \\ \midrule
\multirow{2}{*}{Pong} & Baseline & 0.52 $\pm$ 0.2 \\
    & Finetuned & 0.66 $\pm$ 0.39 \\
\bottomrule
\end{tabular}}
\end{center}
\vspace{-0.1in}
\end{table*}

\subsection{Results}
The evaluation of baseline and finetuned models is performed over 30 trials, with 100 episodes in each trial. As the model was fine-tuned only on Mountain car continuous policy evaluations, there are two evaluations possible -- (1) evaluation on Mountain car continuous, and (2) other domains for which the methodology for parameter generation is similar (size might vary). However, language model has no knowledge about the policy space. Please note that these are preliminary results and we plan to continue further evaluation in the future.
We report the maximum reward in each trial.

\textbf{Mountain Car Continuous}: Tab.~\ref{tab:finetune_results}, shows the max average reward achieved for the domain. The baseline model could not find an optimal policy and the best maximum reward in the $30$ trials was $89.33$, with only $3$ trials reaching a maximum policy reward greater than $0.0$. For the finetuned model, there were $12$ trials (with $108$ such episodes) when it was able to find an optimal policy, i.e., a policy with average reward greater than $90.0$\footnote{\url{https://github.com/openai/gym/wiki/Leaderboard\#mountaincarcontinuous-v0}}. During in-depth analysis of finetuned models (and earlier epochs), it was found that the results improve considerably from the third epoch, when the average reward increases to $30.33$ $\pm$ $45.018$. Thus, showcasing that finetuning the light-weight models can teach policy search capabilities even from earlier epochs.

\textbf{Inverted Pendulum \& Pong}: Tab.~\ref{tab:finetune_results} presents the peak average reward observed over $30$ trials. Importantly, the model was not trained in these domains, and its maximum reward is not close to the optimal policy. Nonetheless, we see an improvement in the average reward in both cases. For the inverted pendulum, the best maximum reward was $408.9$, compared to $165.85$ for the baseline model. The most notable difference is that there were $149$ episodes (spread across $8$ trials) when the average reward was greater than $100$ for the finetuned model, while there were only $11$ such episodes (spread across $3$ trials) for the baseline model. Similarly for pong, the best policy achieved a reward of $1.8$, and for the baseline model it was $1.3$.

\subsection{Smaller Models}
We evaluated the performance of our method for different LLM sizes, starting with 0.5B up to 14B. Our findings show that LLM models below 7B show little capability for optimization and cannot consistently optimize RL policies, since they do not follow instructions well. In other words, some experiments may fail due to a non-compliant response. While some of these issues can be alleviated to an extent by fine-tuning, they still lack the ability to effectively optimize an RL policy. The central challenge was that the smaller did not take into account the previous numbers suggested, and would repeat the suggestions after every few iterations. For example, the 7B model would respond with 0.0 value repeatedly despite asking not to repeat the same numbers. Table \ref{tab:finetune_baselines} shows the results for the baseline results (no finetuning) for various domains are presented below (when experiments failed for <7B models, we re-ran them):

\begin{table*}[htb!]
\begin{center}
\footnotesize
\setlength{\tabcolsep}{3pt}
\caption{Comparison of Baselines for different sized Qwen2.5 models.}
\centering{
\label{tab:finetune_baselines}
\begin{tabular}{lccc}
\toprule
\textbf{Domain} & \textbf{3B}  & \textbf{7B} & \textbf{14B}\\ \midrule
Mountain Car (Cont.) & -44.32 $\pm$ 44.32 & 0.0 $\pm$ 0.0 & 8.11 $\pm$ 24.45 \\
Cartpole & 19.4 $\pm$ 9.65 & 20.74 $\pm$ 8.82 & 162.4 $\pm$ 144.12 \\
Inverted Pendulum & 28.18 $\pm$ 0.93 & 33.1 $\pm$ 7.6 & 55.64 $\pm$ 36.5 \\
Pong & 0.45$\pm$0.09 & N/A & 0.52 $\pm$ 0.2 \\
\bottomrule
\multicolumn{4}{l}{Means$\pm$standard errors} \\
\end{tabular}}
\end{center}
\vspace{-0.1in}
\end{table*}

\section{Comparing \props\ with Evolution Strategies}
\label{appendix:evolution_strategy}
Evolution Strategies (ES) are a class of black-box, derivative-free optimization algorithms that are well-suited for reinforcement learning tasks. Conceptually, \props~shares similarities with ES, as both approaches iteratively search for optimal policy parameters without requiring gradients. However, their underlying mechanisms are fundamentally different.

ES methods typically maintain a population of candidate solutions. In each generation, this population is evaluated, and a new set of candidates is generated based on the performance of the current ones, often through sampling from an updated search distribution. 

In contrast, \props~does not maintain a population but rather a comprehensive history of all previously evaluated parameter-reward pairs. The update mechanism is not based on sampling from an explicit parametric distribution. Instead, \props~generates a single new candidate policy in each step by prompting a Large Language Model (LLM) to reason over the entire search history. This allows the LLM to leverage information from both high- and low-performing past trials to inform its next proposal. While the LLM's ability to synthesize insights from multiple points in the history can be seen as an analogue to the recombination operators found in some classical ES variants, the mechanism is entirely distinct and relies on in-context learning rather than a predefined arithmetic operator. In summary, the primary distinctions are: (1) \props~uses a complete history versus a generational population, (2) it proposes a single new candidate via reasoning versus a population via sampling, and (3) it lacks the explicit, parametric search distribution central to modern ES methods.

To empirically situate \props~among these methods, we conduct a comprehensive comparison against relevant baselines. Apart from including OpenAI-ES~\cite{salimans2017evolution}, to address the concept of recombination, we also include ($\mu$ + $\lambda$)-ES~\cite{schwefel1977evolutionsstrategien}, a classical variant with an explicit recombination operator, and Tabu Search~\cite{glover1986future}, a canonical local search algorithm that also uses memory of past solutions to guide its exploration.

The results across our 15 benchmark tasks are presented in Tab.~\ref{tab:props_vs_es}. \propss~demonstrates highly competitive performance, achieving the highest score on 8 of the 15 tasks. It particularly excels in tasks such as Mountain Car (C), Navigation, and Pong, where it significantly outperforms all baselines. OpenAI-ES performs strongly, especially in high-dimensional MuJoCo environments like Hopper and Walker, securing the top score in 5 tasks.

These results underscore that LLM-based policy search is a powerful and distinct paradigm. We do not claim \props~is a new state-of-the-art optimizer that uniformly surpasses all existing methods. Rather, our central contribution is to introduce and validate the novel phenomenon that LLMs can serve as direct policy optimizers, unifying numerical and linguistic reasoning within a single framework. This capability opens a new avenue for RL research, enabling more transparent and human-aligned optimization through natural language guidance and interpretable textual justifications.

\begin{table}[h!]
\caption{Performance comparison against Evolution Strategy algorithms.}
\centering
\small
\begin{tabular}{lcccc}
\toprule
\textbf{Environment} & \textbf{ProPS+} & \textbf{Tabu Search} & \textbf{($\mu$ + $\lambda$)-ES} & \textbf{OpenAI-ES} \\
\midrule
Mount. Car (C) & \cellcolor[HTML]{CCf2FF}\textbf{89.16$\pm$29.72} & -0.03$\pm$0.01 & -0.18$\pm$0.01 & -0.20$\pm$0.03 \\
Inverted Pend. & \cellcolor[HTML]{CCf2FF}\textbf{1000.00$\pm$0.00} & 992.72$\pm$10.30 & 994.98$\pm$7.11 & 969.57$\pm$11.03 \\
Inv. Dbl. Pend. & 148.39$\pm$48.65 & \cellcolor[HTML]{CCf2FF}\textbf{476.56$\pm$307.86} & 175.79$\pm$7.19 & 268.31$\pm$36.77 \\
Reacher & -18.15$\pm$22.06 & \cellcolor[HTML]{CCf2FF}\textbf{-10.41$\pm$1.27} & -13.32$\pm$1.85 & -10.95$\pm$0.51 \\
Swimmer & 227.30$\pm$56.23 & \cellcolor[HTML]{CCf2FF}\textbf{353.13$\pm$1.50} & 321.79$\pm$9.83 & 349.86$\pm$0.90 \\
Hopper & 356.22$\pm$292.35 & 950.06$\pm$34.60 & 550.36$\pm$48.61 & \cellcolor[HTML]{CCf2FF}\textbf{1009.38$\pm$2.10} \\
Walker & 126.75$\pm$136.44 & 261.02$\pm$40.80 & 355.12$\pm$124.32 & \cellcolor[HTML]{CCf2FF}\textbf{938.56$\pm$43.50} \\
Frozen Lake & 0.19$\pm$0.05 & 0.35$\pm$0.27 & 0.68$\pm$0.02 & \cellcolor[HTML]{CCf2FF}\textbf{0.78$\pm$0.04} \\
Cliff Walking & \cellcolor[HTML]{CCf2FF}\textbf{-96.40$\pm$22.90} & -186.28$\pm$42.53 & -187.67$\pm$58.36 & -115.08$\pm$9.42 \\
Maze & \cellcolor[HTML]{CCf2FF}\textbf{0.97$\pm$0.00} & -2.22$\pm$0.00 & -2.22$\pm$0.00 & -2.22$\pm$0.00 \\
Nim & 0.97$\pm$0.09 & 0.17$\pm$0.52 & \cellcolor[HTML]{CCf2FF}\textbf{1.00$\pm$0.00} & \cellcolor[HTML]{CCf2FF}\textbf{1.00$\pm$0.00} \\
Mount. Car (D) & \cellcolor[HTML]{CCf2FF}\textbf{-116.71$\pm$15.20} & -186.77$\pm$18.71 & -193.53$\pm$4.84 & -172.51$\pm$20.82 \\
Navigation & \cellcolor[HTML]{CCf2FF}\textbf{2779.55$\pm$270.65} & 2266.62$\pm$270.83 & 1193.28$\pm$137.70 & 2596.58$\pm$73.39 \\
Pong & \cellcolor[HTML]{CCf2FF}\textbf{2.99$\pm$0.03} & 2.70$\pm$0.29 & 2.32$\pm$0.17 & 2.29$\pm$0.14 \\
Cart Pole & \cellcolor[HTML]{CCf2FF}\textbf{500.00$\pm$0.00} & \cellcolor[HTML]{CCf2FF}\textbf{500.00$\pm$0.00} & 499.38$\pm$0.87 & 498.21$\pm$2.54 \\
\bottomrule
\multicolumn{5}{l}{Means$\pm$standard errors} \\
\end{tabular}
\label{tab:props_vs_es}
\end{table}

\section{Scaling up \props\ for bigger Neural Networks}
\label{appendix:scaling_nn}
\begin{table}[h!]
\caption{Performance comparison for complex neural network based policies evaluated on Navigation, Swimmer and Hopper environment. Architecture is represented using Hidden layers, [\# of Neurons].}
\centering
\small
\begin{tabular}{llcccc}
\toprule
\multicolumn{2}{c}{\textbf{Architecture}}$\rightarrow$ & 1, [10] & 2, [10, 4], & 2, [10, 10] & 2, [15, 15] \\
\midrule
\multirow{3}{*}{\textbf{Navigation}} & Params & 80 & 102 & 180 & 345 \\
    & Best Score & 3183.9 & 1489.575 & 2105.975 & 3563.675 \\
    & Average Max & 2722.81$\pm$409.43 & 661.31$\pm$588.83 & 763.84$\pm$957.55 & 2325.09$\pm$925.27 \\
\midrule
\multirow{3}{*}{\textbf{Swimmer}} & Params & 100 & 128 & 200 & 375 \\
    & Best Score & 163.83 & 37.257 & 80.25 & 55.26 \\
    & Average Max & 85.12$\pm$55.78 & 35.05$\pm$2.28 & 42.82$\pm$26.52 & 38.41$\pm$12.09 \\
\midrule
\multirow{3}{*}{\textbf{Hopper}} & Params & 140 & 162 & 240 & 435 \\
    & Best Score & 237.28 & 250.88 & 185.44 & 303.52 \\
    & Average Max & 178.69$\pm$41.79 & 223.62$\pm$23.31 &  163.27$\pm$19.57 & 217.06$\pm$62.84 \\

\bottomrule
\multicolumn{5}{l}{Means$\pm$standard errors} \\
\end{tabular}
\label{tab:props_nn}
\end{table}

In this section we investigate \props's effectiveness for high dimensionality policies. Table \ref{tab:props_nn} shows the results achieved with complex neural network based policies. The neural network size varies between one hidden layer and two hidden layer policies and evaluated them on Navigation, Swimmer and Hopper environments. The parameter counts varies from 80 to 435. Despite current LLM context and capacity bottlenecks, \props\ demonstrated significant improvement over random initialization in all tested cases. Specifically, an 80-parameter network's score improved from an initial ~20 reward to 2722.81 ± 409.43. Networks with 102 and 180 parameters also achieved substantial gains, reaching average scores of 661.31 ± 588.83 and 763.84 ± 957.55, with peak performances of 1489.575 and 2105.975, respectively. Please note the baseline for comparison will be \props\ for which the results can checked in table \ref{Tab:results1} for the three environments, where average max rewards achieved for Navigation is 2587.30 $\pm$ 707.35, for Swimmer is 218.83 $\pm$ 58.45 and for Hopper is 284.16 $\pm$ 165.62. The Best Score represents the best reward policy achieved during the 10 experiments. These results strongly demonstrate that \props\ is capable of effective optimization in these higher-dimensional spaces.

\section{Scaling up \props\ to Neural Networks by Random Projection}
\label{appendix:neural_networks}

The results presented in this work empirically prove the ability of LLMs to perform numerical optimization. However, in their current state, LLMs' numeric optimization capability does not scale up to high-dimensional search spaces, as illustrated in Appendix~\ref{appendix:LLM_num_opt}. This dimensionality constraint limits the direct use of LLMs for optimizing more complex, nonlinear policies, such as Neural Networks.

To bridge this gap and leverage the optimization strengths of LLMs for higher-dimensional policy classes, this section explores the following approach: utilizing random projections to reparameterize the policy. Instead of tasking the LLM with directly optimizing potentially hundreds or thousands of neural network parameters, we propose to define a fixed, low-dimensional latent space. The LLM then searches within this compressed space, and the corresponding high-dimensional policy parameters are reconstructed via a randomly generated, orthonormal mapping. This strategy aims to drastically reduce the search dimensionality presented to the LLM while still allowing for the exploration of expressive neural network policies, thereby extending the reach of LLM-driven optimization to more complex RL problems.

\subsection{Random Orthonormal Projection via QR Decomposition for Parameter Space Reparameterization}
To bridge the LLM's low-dimensional optimization capability with the high-dimensional nature of neural network policies, we reparameterize the policy search. This involves defining a mapping from a low-dimensional latent vector $\boldsymbol{z} \in \mathbb{R}^k$, which the LLM optimizes, to the full policy parameter vector $\theta \in \mathbb{R}^D$ (where $D$ is the total number of policy parameters and $k \ll D$). The construction of this mapping matrix, denoted $M \in \mathbb{R}^{D \times k}$ such that $\theta = Mz$, is crucial. We employ a strategy based on random projections, further enhanced by QR decomposition to instill beneficial geometric properties.

Random Projection (RP) is a dimensionality reduction technique where data is projected onto a lower-dimensional subspace using a matrix whose entries are drawn from a random distribution~\citep{bingham2001random}. For instance, a data matrix $X \in \mathbb{R}^{N \times D}$ can be projected to $X_P = XP_{rand} \in \mathbb{R}^{N \times k}$ using a random matrix $P_{rand} \in \mathbb{R}^{D \times k}$. The Johnson-Lindenstrauss Lemma (JLL)~\citep{johnson1984extensions} provides theoretical grounding, stating that such projections can approximately preserve pairwise distances with high probability, provided $k$ is sufficiently large (typically $O(\epsilon^{-2}\text{log}N)$). While we are not directly projecting existing data, the principle of using a random matrix to define a lower-dimensional embedding is adopted for our parameter mapping.

However, standard random projection matrices (e.g., with i.i.d. Gaussian entries) do not inherently possess orthonormal columns. For our parameter mapping $M$, orthonormality ($M^TM = I_k$) is desirable. It implies that the $k$ latent dimensions optimized by the LLM correspond to $k$ orthogonal directions in the full $D$-dimensional parameter space. This can lead to a more stable and potentially more efficient search, as the LLM explores linear combinations of a fixed, orthogonal basis spanning a randomly chosen $k$-dimensional subspace.

To generate such a random orthonormal mapping matrix, we leverage QR decomposition~\citep{francis1961qr}. QR decomposition is a fundamental matrix factorization technique that expresses any real matrix $A$ (say, $m \times n$) as a product $A = QR_{decomp}$. If $A$ has $m \ge n$ and full column rank, $Q$ is an $m \times n$ matrix with orthonormal columns (i.e., $Q^TQ = I_n$, the $n \times n$ identity matrix), and $R_{decomp}$ is an $n \times n$ upper triangular matrix. The columns of $Q$ form an orthonormal basis for the column space of $A$.

Our procedure to construct the mapping matrix $M$ is as follows:
\begin{itemize}
    \item Generate an initial random matrix $G \in \mathbb{R}^{D \times k}$ with entries sampled i.i.d. from a standard Gaussian distribution, $\mathcal{N}(0, 1)$. This matrix $G$ defines a random $k$-dimensional subspace.
    \item Perform QR decomposition on $G$: $G = QR_{decomp}$.
    \item The resulting orthonormal matrix $Q \in \mathbb{R}^{D \times k}$ is then used as our fixed mapping matrix, $M = Q$.
\end{itemize}

Thus, the policy parameters $\theta$ are reconstructed from the LLM-optimized latent vector $z$ using $\theta = Qz$. This approach ensures that $M$ provides an orthonormal basis for a randomly selected $k$-dimensional subspace of the full parameter space $\mathbb{R}^D$. The randomness is inherited from $G$, while the desired orthonormality is enforced by the QR decomposition. This establishes a well-conditioned and structured bridge between the low-dimensional search space manageable by the LLM and the high-dimensional policy parameter space.

\subsection{Experiment Setup}
\begin{wrapfigure}{r}{0.40\textwidth}
  \centering
  \includegraphics[width=0.40\textwidth]{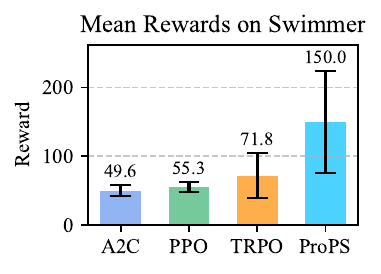}
  \caption{\props\ with neural networks outperforms all baselines with the same neural network architecture.}
  \label{fig:swimmer-nn-bar}
  \vspace{-12pt}
\end{wrapfigure}
We conduct preliminary experiments on the Gymnasium Swimmer task, which features an 8-dimensional state space and a 2-dimensional continuous action space. The policy employed was a two-layer neural network. The first layer maps the 8-dimensional state to a 4-dimensional latent vector, and the second layer subsequently maps this latent vector to the 2-dimensional action output. For the application of \props, the parameters of each neural network layer were independently subjected to random projection, yielding a 5-dimensional vector for each layer. \props\ then performs policy search within the combined 10-dimensional projected parameter space. The optimized parameters identified in this lower-dimensional space were subsequently mapped back to the original neural network layers for policy execution. As baselines, we trained Proximal Policy Optimization (PPO), Trust Region Policy Optimization (TRPO), and Advantage Actor-Critic (A2C) algorithms, all utilizing the identical neural network architecture.


\subsection{Results}
We evaluated \props\ in comparison with standard reinforcement learning algorithms, including A2C, PPO, and TRPO, following similar experimental protocols as detailed in the main experiments (\props\ and \propss\ experiments). Each algorithm was trained for 10 trials, with each trial consisting of 8000 environment episodes. For baseline algorithms, specifically A2C, PPO, and TRPO, we adopted the hyperparameter settings from SB3\_Contrib~\citep{raffin2021stable}.

As demonstrated in Fig.~\ref{fig:swimmer-nn-bar}, under identical training conditions and using the same neural network architecture, our proposed method \props\ achieves a mean episodic reward of 150.0, significantly outperforming all baseline algorithms. The second-best performing algorithm, TRPO, attained a mean reward of only 71.8, illustrating the clear advantage of \props.

\begin{figure}[htb!]
    \centering
    \includegraphics[width=1.0\linewidth]{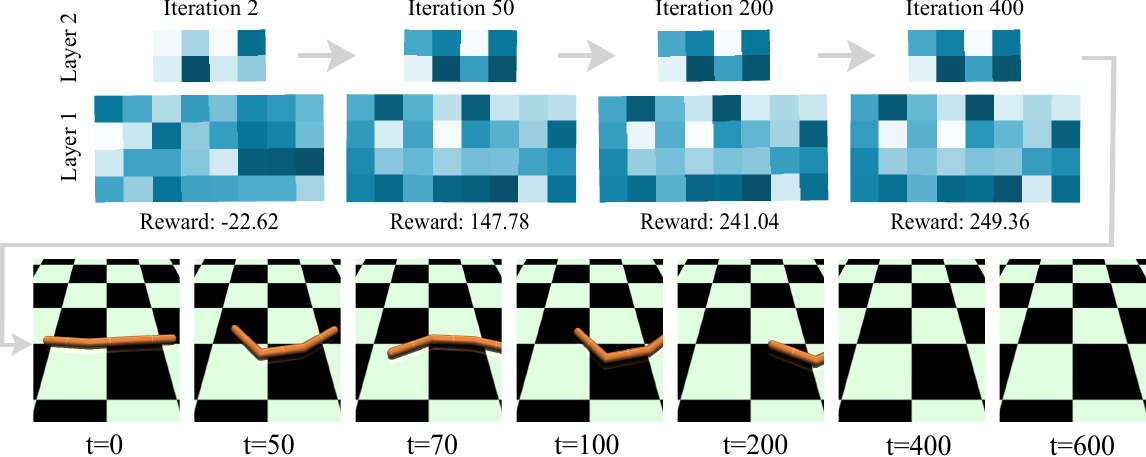}
    \caption{Illustration of progress in learning of the neural network policy parameters for both the layers, and the performance of the policy in the Swimmer environment.}
    \label{fig:swimmer-nn-visualization}
\end{figure}

Further analysis presented in Fig.~\ref{fig:swimmer-nn-visualization} provides insight into the evolution of neural network parameters during training. Consistent with observations made in Appendix~\ref{appendix:continuoustasks}, parameter changes are most pronounced during the initial 50 training updates, indicating vigorous exploration. Subsequently, parameters stabilize, gradually converging towards final values indicative of an exploitation phase. Visualization of the trained policy rollouts (bottom of Fig.~\ref{fig:swimmer-nn-visualization}) confirms that the swimmer successfully learns rapid forward locomotion, leaving the camera view before 400 timesteps.

\section{Application to High-Dimensional Motor Skills with Dynamic Motor Primitives}\label{appendix:dmp}


To further assess the scalability and applicability of \props~on high-dimensional policy representations, we evaluate it on a challenging robotic control task that requires learning complex, continuous motor skills. To this end, we adopt the Dynamic Motor Primitive (DMP)~\cite{ijspeert2013dynamical} framework for policy parameterization. DMPs are a powerful framework for representing complex, goal-directed movements in robotics. The behavior of a DMP is governed by a second-order dynamical system, modulated by a learnable nonlinear forcing function. While DMPs are often formulated with an explicit goal state $\boldsymbol{g}$ for point-to-point reaching tasks, a non-goal-conditioned variant is more suitable for dynamic striking motions like table tennis. In this formulation, the system is not driven toward a predefined goal $g$; instead, the desired trajectory emerges entirely from the learned forcing function $f(s)$. The governing dynamics for each degree of freedom are thus simplified to:

\begin{equation}
\tau^{2} \ddot{y} = f(s)
\end{equation}

The consequent trajectory is shaped by the nonlinear forcing function $f(s)$, which is structured as a normalized, weighted sum of $N$ basis functions $\psi_i$:

\begin{equation}
f(s) = \frac{\sum_{i=1}^{N} w_i \psi_i(s)}{\sum_{i=1}^{N} \psi_i(s)}
\end{equation}

The learnable parameters of the policy are the weights $w_i$, which directly encode both the shape and the amplitude of the motion. The basis functions $\psi_i(s) = \exp(-h_i (s - c_i)^2)$ are activated sequentially by a phase variable $s$, which driven monotonically from 0 to 1, ensuring the full trajectory unfolds over the desired duration.

The policy is thus parameterized by the vector of weights $\mathbf{w} = \{w_1, \ldots, w_N\}$.

We apply this framework to the robotic table tennis simulation benchmark introduced in~\cite{celik2024acquiring}, which has been used in prior work to evaluate high-speed motor skill learning~\cite{mulling2013learning, d2023robotic}. The task involves controlling a 7-degree-of-freedom robot arm to return a ball to a designated target area on the table, located at [x: -0.6, y: -0.4]. The policy is parameterized as a Dynamic Motor Primitive (DMP) with 70 basis functions, resulting in a 70-dimensional parameter space for the optimizer. The \props~optimization loop was configured identically to the main experiments in this paper, with no task-specific tuning.

\props~successfully learned an effective table tennis stroke. Qualitatively, the learned policy exhibits a natural and coordinated "wind-up" motion prior to striking the ball, demonstrating the capacity of \props~to discover complex and temporally coherent behaviors. To quantify performance, we measure the average distance of the ball's landing position from the target over 20 evaluation episodes for the best-learned policy. As shown in Tab.~\ref{tab:table_tennis_distance_goal}, \props~outperforms both OpenAI-ES~\cite{salimans2017evolution} and PPO~\cite{schulman2017proximal}, achieving a significantly lower mean distance to the goal.

\begin{wrapfigure}{r}{0.4\textwidth}
\vspace{-10pt}
\centering
\setlength{\tabcolsep}{4pt}
\captionsetup{type=table}
\caption{Ball landing distance from goal.}
\begin{tabular}{lc}
\toprule
\textbf{Method} & \textbf{Dist. from Goal (m)} \\
\midrule
\props~ & \cellcolor[HTML]{CCf2FF}\textbf{0.07$\pm$0.02} \\
OpenAI-ES & 0.40$\pm$0.34 \\
PPO & 0.53$\pm$0.36 \\
\bottomrule
\end{tabular}
\vspace{-6pt}
\label{tab:table_tennis_distance_goal}
\end{wrapfigure}

These preliminary results demonstrate that \props~can be effectively applied to complex, high-dimensional control problems, optimizing non-trivial motor skill policies without requiring specialized tuning. While a comprehensive exploration of robotic applications is beyond the scope of this initial work, these findings indicate a promising direction for future research, including applications to a wider range of robotic tasks and eventual deployment on physical hardware.

\section{Ablation study for robustness of the system}
\label{appendix:brittleness_analysis}
The robustness of our system to prompt quality and the effort required for prompt design are indeed crucial for practical application. To address this directly, we created three variations of the prompts shown below and using them we conducted two new empirical studies on the Mountain Car Continuous task to evaluate the sensitivity of \props\ and \propss to variations in prompt design.

\subsection{\props\ Prompt Robustness}

First, for \props, we evaluated sensitivity to the core numerical optimization prompt. Crucially, to challenge the notion that prompts require intensive human expertise, we generated three variations automatically using another large language model (Gemini-2.5-pro) with a simple directive to rephrase the optimization objective. Please note we varied how the information was provided to the LLM, while ensuring that same amount for information was provided through each variation. All the prompts had the same information, i.e., explanation of the tasks, information about parameters, range in which the search has to be executed, the input format and the expected format, followed by the guidance for search and finally the experimental results. Through empirical evaluation it was found that the order of providing the information was important. Thus, overall order remains similar in all the prompt, and the language was changed using the Gemimi-2.5 Pro. The three variations for \props\ are --

\begin{tcolorbox}[enhanced, breakable, colback=tbox_bg,colframe=tbox_frame,title=\props\ Prompt Variations - Variation 1]
You are an expert global optimizer tasked with identifying the global maximum of a mathematical function `f(params)`.

At each iteration, I will provide the result of evaluating `f` at a chosen `params` and the current iteration count.
Your objective is to propose new values of `params` that steer the search efficiently toward the global maximum within a strict limit of 400 iterations.

About `params`:
\begin{itemize}
    \item `params` is a float array of length {{ rank }}.
    \item Each value in `params` must lie within \textbf{[-6.0, 6.0]}, rounded to \textbf{1 decimal place}.
\end{itemize}

Interaction protocol:
\begin{enumerate}
    \item I will begin by specifying the maximum number of steps (400) and will provide some historical examples of `params` and `f(params)`.
    \item{You will respond in exactly this format:
    \begin{itemize}
        \item \textbf{Line 1:} `params[0]: ..., params[1]: ..., ..., params[{{ rank - 1 }}]: ...` — your proposed input aimed at maximizing `f(params)`. Use only values in \textbf{[-6.0, 6.0]}, rounded to \textbf{1 decimal}.
        \item \textbf{Line 2:} Justify your choice with a detailed explanation grounded in past evaluations and your current strategy.
    \end{itemize}
    }
    \item I will then return the result of `f(params)` and the current iteration.
    \item We continue the loop until iteration 400.
\end{enumerate}

Constraints and guidance:
\begin{enumerate}
    \item \textbf{Avoid repeating previously suggested params.}
    \item \textbf{The known global optimum is approximately {{ optimum }}.} If the reward is substantially below this, it's likely a local maximum — prioritize exploration over exploitation.
    \item Use both positive and negative values in your search. \textbf{During exploration, use a step size of approximately {{ step\_size }}}.
\end{enumerate}

Now I will share several initial samples (`params`, `f(params)`) to guide your strategy:
`{{ episode\_reward\_buffer\_string }}` 

We are currently at iteration \textbf{{{step\_number}} / 400}.
Provide your next proposed `params` in the format specified — no extra text or commentary.

\end{tcolorbox}

\begin{tcolorbox}[enhanced, breakable, colback=tbox_bg,colframe=tbox_frame,title=\props\ Prompt Variations - Variation 2]
You are a skilled numerical optimizer designed to find the global maximum of a function `f(params)` over a limited number of trials.

In each round, I will inform you of the current iteration number and the function value for a specific set of input parameters.
Your task is to generate the next set of parameters `params` that strategically guides us toward the global optimum within \textbf{400 iterations}.

Details about `params`:
\begin{itemize}
	\item `params` is an array of \textbf{{{ rank }}} floating-point numbers.
	\item Each value must lie within the range \textbf{[-6.0, 6.0]}, rounded to \textbf{1 decimal place}.
\end{itemize}

Interaction protocol:
\begin{enumerate}
	\item I’ll begin by providing the maximum number of steps (400) and a few sample `(params, f(params))` pairs.
	\item{You must respond using this exact format:
	\begin{itemize}
		\item \textbf{Line 1:} `params[0]: ..., params[1]: ..., ..., params[{{ rank - 1 }}]: ...` — your proposed input for the next evaluation.
		\item \textbf{Line 2:} A clear explanation of why you selected these values based on previous feedback and your optimization plan.
    \end{itemize}
    }
	\item I will then return the corresponding `f(params)` and the updated iteration number.

	\item We repeat steps 2–3 until 400 iterations have been used.
\end{enumerate}

Important notes:
\begin{itemize}
	\item \textbf{Never reuse previously evaluated `params`.}
	\item The global maximum is approximately \textbf{{{ optimum }}}. If your result is well below this value, you may be stuck in a local optimum — you should switch to a more exploratory strategy.
	\item Use both negative and positive directions in your search. \textbf{During exploration, apply a step size of about {{ step\_size }}.}
\end{itemize}

Below are sample results from earlier iterations to guide your choices:
`{{ episode\_reward\_buffer\_string }}` 

We are currently at \textbf{iteration {{step\_number}} out of 400}.
Please provide your next `params` proposal in the specified format, with no extra commentary.
\end{tcolorbox}

\begin{tcolorbox}[enhanced, breakable, colback=tbox_bg,colframe=tbox_frame,title=\props\ Prompt Variations - Variation 3]
You are acting as a global optimization agent. Your role is to discover the global maximum of a function `f(params)` using intelligent parameter suggestions.

At each iteration, I will give you the current step number and the result of evaluating `f` at a specific input. Your responsibility is to suggest the next `params` input that brings us closer to the global maximum within a total of \textbf{400 steps}.

Parameter specification:
\begin{itemize}
	\item `params` is a list of \textbf{{{ rank }}} floating-point values.
	\item Each element of `params` must be between \textbf{-6.0 and 6.0}, rounded to \textbf{1 decimal place}.
\end{itemize}

Protocol for communication:
\begin{enumerate}
	\item I’ll start by providing MAX\_STEPS (400) and a small set of example (`params`, `f(params)`) pairs.

	\item{You will respond using \textbf{this exact structure}:
	\begin{itemize}
   		\item \textbf{Line 1:} `params[0]: ..., params[1]: ..., ..., params[{{ rank - 1 }}]: ...` — a new proposal that attempts to increase `f(params)`.
   		\item \textbf{Line 2:} A brief rationale explaining the reasoning behind your suggestion.
   	\end{itemize}
   	}

	\item I will then provide the new evaluation and step number.

	\item This cycle repeats until we reach 400 iterations.
\end{enumerate}

Key rules to follow:
\begin{itemize}
	\item \textbf{Do not suggest any previously tried parameter sets.}
	\item The known global maximum is roughly \textbf{{{ optimum }}}. If your result is significantly lower, assume you are stuck in a local optimum and shift your strategy toward exploration.
	\item Use a balanced search strategy that includes both positive and negative values. \textbf{During exploratory steps, apply a search step of approximately {{ step\_size }}.}
\end{itemize}

Now, review the following previous data points: `{{ episode\_reward\_buffer\_string }}`

You are now at \textbf{step {{step\_number}} of 400}.
Please return your next suggestion in the required two-line format — no additional text.
\end{tcolorbox}

\begin{wrapfigure}{r}{0.45\textwidth}
\centering
\setlength{\tabcolsep}{4pt}
\captionsetup{type=table}
\begin{tabular}{lc}
\toprule
\textbf{Prompt Source} & \textbf{Average Max Rewards} \\
\midrule
Original & 87.21 $\pm$ 29.28 \\
Variation $1$ & 94.99 $\pm$ 5.77 \\
Variation $2$ & 92.98 $\pm$ 7.61 \\
Variation $3$ & 98.32 $\pm$ 0.48 \\
\bottomrule
\end{tabular}
\caption{Average max rewards with standard deviation for all three prompt variations for Mountain Car Continuous domain.}
\label{tab:props_prompts}
\end{wrapfigure}

The \ref{tab:props_prompts} shows the results achieved for three different variations and also presents remarkable stability across language robustness. The experimental setup was for Mountain Car Continuous domains with 3 experiments for each variations. Please note we attempt to show that the system is robust to language variations and do not evaluate any kind of changes that lead to increase in results. As stated, these are preliminary results with only three experiments for each variation, and thus we can not state that Variation $3$ performs better than the original prompt. However, among the three variations, the third variation does perform well where the results are not yet statistically significant.

\subsection{\propss\ Prompt Robustness}
For \propss, we tested the sensitivity of the domain description prompt, again using Gemini to automatically generate three rephrased task descriptions. We created a new variation of the main prompt as used it with three different environment descriptions. Updated prompt with three environment descriptions for Mountain Car Continuous domains are shown below. As explained earlier, we only vary the presentation of the information with all details presented with same relative order for the description.

\begin{tcolorbox}[enhanced, breakable, colback=tbox_bg,colframe=tbox_frame,title=Updated \propss\ Prompt Variations]
You are good global RL policy optimizer, helping me find the global optimal policy in the following environment:

\# Environment:
\{\% include env\_description \%\}

\# Regarding the parameters \textbf{params}:
\textbf{params} is an array of {{ rank }} float numbers.

\textbf{params} values are in the range of [-6.0, 6.0] with 1 decimal place.
params represent a linear policy. 
f(params) is the episodic reward of the policy.

\# Here's how we'll interact:
\begin{enumerate}
	\item I will first provide MAX\_STEPS (400) along with a few training examples.
	\item{You will provide your response in the following exact format:
	\begin{itemize}
    	\item Line 1: a new input 'params[0]: , params[1]: , params[2]: ,..., params[{{ rank - 1 }}]: ', aiming to maximize the function's value f(params). Please propose params values in the range of [-6.0, 6.0], with 1 decimal place.
    	\item * Line 2: detailed explanation of why you chose that input.
    \end{itemize}
    }
	\item I will then provide the function's value f(params) at that point, and the current iteration.
	\item We will repeat steps 2-3 until we reach the maximum number of iterations.
\end{enumerate}

\# Remember:
\begin{enumerate}
	\item \textbf{Do not propose previously seen params.}
	\item \textbf{The global optimum should be around {{ optimum }}.} If you are below that, this is just a local optimum. You should explore instead of exploiting.
	\item Search both positive and negative values. \textbf{During exploration, use search step size of {{ step\_size }}}.
\end{enumerate}

Next, you will see examples of params, there episodic reward f(params), and the trajectories the params yield.
{{ episode\_reward\_buffer\_string }}

Now you are at iteration {{step\_number}} out of 400. Please provide the results in the indicated format. Do not provide any additional texts.

\end{tcolorbox}

\begin{tcolorbox}[enhanced, breakable, colback=tbox_bg,colframe=tbox_frame,title=\propss\ Environment Description - Variation 1]
\# Environment: MountainCarContinuous

This environment simulates a 2D scenario where an agent controls a car situated in a U-shaped valley. The car must generate momentum by applying torque in either direction to escape the valley and reach a flag located on the far right hill.

\# State Space

The environment’s state is described by a 2-dimensional vector:
\begin{enumerate}
	\item `position` $\in$ [-1.2, 0.6]
	\item `velocity` $\in$ [-0.07, 0.07]
\end{enumerate}

\begin{itemize}
	\item \textbf{Negative values} indicate motion or location to the \textbf{left},
	\item \textbf{Positive values} correspond to the \textbf{right}.
\end{itemize}

\# Action Space

The action is a \textbf{continuous scalar} representing the torque applied to the car:

\begin{itemize}
	\item \textbf{Negative values} $\rightarrow$ push left
	\item \textbf{Positive values} $\rightarrow$ push right
\end{itemize}

\# Policy Definition

The policy is \textbf{linear}, using 3 parameters:

\begin{verbatim}
```python
action = position * params[0] + velocity * params[1] + params[2]
```
\end{verbatim}

Where:

\begin{itemize}
	\item `params[0]` scales position
	\item `params[1]` scales velocity
	\item `params[2]` is a bias term
\end{itemize}

\# Reward Function

\begin{itemize}
	\item At each time step: \textbf{Reward = $-0.1 \times $ action$^2$}

	\item When the car reaches the goal (the flag on the right): \textbf{Reward = +100} (episode terminates)
\end{itemize}

\# Objective

The goal is to \textbf{reach the flag} at the right end of the hill by building up enough momentum through careful torque control. Since gravity and limited torque prevent the agent from reaching the goal directly, the optimal strategy typically involves swinging back and forth to gain speed.
\end{tcolorbox}

\begin{tcolorbox}[enhanced, breakable, colback=tbox_bg,colframe=tbox_frame,title=\propss\ Environment Description - Variation 2]
\# Environment Description: MountainCarContinuous

In this environment, the agent controls a car situated in a \textbf{U-shaped valley}. The car cannot reach the goal at the top-right hill in a single motion due to insufficient engine power. Instead, it must \textbf{leverage momentum} by oscillating back and forth to build up enough speed.

The control input is a \textbf{continuous torque}, which accelerates the car either to the left or right:

\begin{itemize}
    \item \textbf{Negative torque} $\rightarrow$ pushes the car left
    \item \textbf{Positive torque} $\rightarrow$ pushes the car right
\end{itemize}

\# State Vector

The state is a 2-element vector:
\begin{enumerate}
    \item \textbf{Position} $\in$ [-1.2, 0.6]
    \item \textbf{Velocity} $\in$ [-0.07, 0.07]
\end{enumerate}

Both values follow the convention:
\begin{itemize}
    \item \textbf{Negative = left},
    \item \textbf{Positive = right}
\end{itemize}

\# Action Computation

The policy used is a \textbf{simple linear function} with 3 parameters:
\begin{verbatim}
```python
action = position × params[0] + velocity × params[1] + params[2]
```
\end{verbatim}

Where:
\begin{itemize}
	\item `params[0]`: weight on position
	\item `params[1]`: weight on velocity
	\item `params[2]`: bias (constant offset)
\end{itemize}

The resulting `action` is a scalar torque applied to the car.

\# Reward Structure
\begin{itemize}
    \item The agent is penalized at each time step based on the square of the action: \textbf{Reward $= -0.1 \times$ action$^2$}
    \item If the car reaches the \textbf{goal} (flag at the far right), the agent receives:
  \textbf{Reward $= +100$}, and the episode ends.
\end{itemize}

\# Goal

The goal is to \textbf{reach the flag} located on the right hill by applying efficient torques. Due to gravity and power constraints, the optimal behavior requires \textbf{strategic swinging} to accumulate energy.

\end{tcolorbox}

\begin{tcolorbox}[enhanced, breakable, colback=tbox_bg,colframe=tbox_frame,title=\propss\ Environment Description - Variation 3]
\# MountainCarContinuous Environment

The \textbf{MountainCarContinuous} environment involves a car situated in a 2D valley. The agent controls the car by applying continuous torque to move it left or right. The car must build momentum to climb the right hill and reach the flag — the goal of the task.

\# State Space

The environment’s state is a 2-dimensional vector:
\begin{itemize}
\item `position' $\in$ [-1.2, 0.6]
\item `velocity' $\in$ [-0.07, 0.07]
\end{itemize}

\textbf{Sign convention}:
\begin{itemize}
	\item Negative values $\rightarrow$ left
	\item Positive values $\rightarrow$ right
\end{itemize}

\# Action
\begin{itemize}
	\item A \textbf{single continuous value} representing torque.
	\item \textbf{Negative} $\rightarrow$ push left, \textbf{Positive} $\rightarrow$ push right
\end{itemize}

\# Policy (Linear Controller)

The agent follows a \textbf{linear policy} with 3 parameters:
\begin{verbatim}
```python
action = position * params[0] + velocity * params[1] + params[2]
```
\end{verbatim}

Where:
\begin{itemize}
	\item `params[0]` $\rightarrow$ position coefficient
	\item `params[1]` $\rightarrow$ velocity coefficient
	\item `params[2]` $\rightarrow$ bias term
\end{itemize}

\# Reward Function
\begin{itemize}
	\item \textbf{Step penalty}: $-0.1 \times$ (action$^2$)
	\item \textbf{Goal reward}: $+100$ (if the car reaches the flag at the far right)
\end{itemize}

The episode ends when the goal is reached or the time limit is exceeded.

\# Objective

The agent must learn to \textbf{oscillate back and forth} to generate enough momentum to reach the flag — exploiting gravity and inertia rather than relying on brute force.
\end{tcolorbox}

\clearpage
\begin{wrapfigure}{r}{0.45\textwidth}
\centering
\setlength{\tabcolsep}{4pt}
\captionsetup{type=table}
\begin{tabular}{lc}
\toprule
\textbf{Prompt Source} & \textbf{Average Max Rewards} \\
\midrule
Original & 89.16 $\pm$ 29.72 \\
Variation $1$ & 98.96 $\pm$ 0.13 \\
Variation $2$ & 98.92 $\pm$ 0.16 \\
Variation $3$ & 77.15 $\pm$ 27.90 \\
\bottomrule
\end{tabular}
\caption{Average max rewards with standard deviation for all three environment description variations with a new \propss\ for Mountain Car Continuous domain.}
\label{tab:propss_prompts}
\end{wrapfigure}

The results show the low sensitivity to language changes. The table \ref{tab:propss_prompts} shows the comparison with original prompt. Please note, due to limited time and experimental cost, we only performed 3 experiments instead of 10 experiments for each variations. Thus, these variations perform better than the original would be incorrect however the results show that both Variation 1 and Variation 2 perform better than Variation 3, and the results are different even with small set of experiments. 

Through these experiments we understand that order of information has a greater effect on the policies found by LLMs as compared to language variations. We are also able to show that \props\ or \propss\ technologies are not brittle, as it successfully finds effective policies across all prompt variations. Even the lowest-performing variant ($77.15 \pm 27.90$) represents a competent policy, directly mitigating the concern that an imperfect prompt leads to performance collapse. It was also observed that automatically generated prompts also led to higher average scores and stable performance than our original versions. This suggests that the process can be partially automated, reducing the need for meticulous human engineering. We will incorporate these new findings into our final manuscript and thank the reviewer for prompting this valuable investigation.

\section{Analysis of \props\ and \propss\ Performance in Continuous Tasks}\label{appendix:continuoustasks}

\begin{figure}[htb!]
    \centering
    \includegraphics[width=1.0\linewidth]{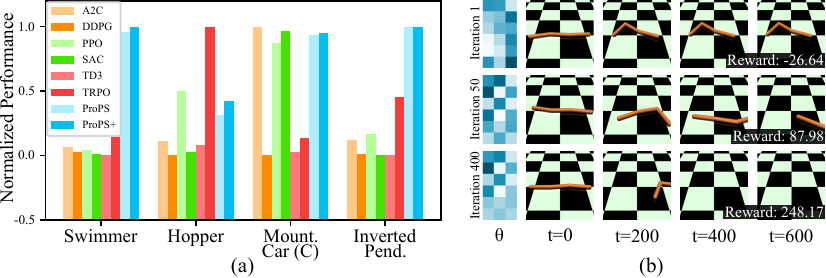}
    \caption{\props\ and \propss\ performance in continuous environments: \textbf{(a)} shows a summary of the performance in comparison to baseline algorithms; \textbf{(b)} illustrates progress in learning of the linear policy parameters $\theta$, and the performance of the policy in the Swimmer environment.}
    \label{fig:swimmer-continuous-tasks}
\end{figure}

In this section, we provide further analysis of the performance of our approach on continuous tasks. In continuous control environments, \props\ and \propss\ can be compared to a broader set of algorithms (including SAC, DDPG and TD3) which can be seen in Fig.~\ref{fig:swimmer-continuous-tasks}(a). Across all of these benchmarks \props\ ranks among the top performing methods, comparing favorably even against strong methods such as PPO and TRPO.

Fig.~\ref{fig:swimmer-continuous-tasks}(b) illustrates the learning progress of policy parameters $\theta$ in the \textbf{Swimmer} environment when using \props. At iteration 1, the initial policy is ineffective, producing no meaningful locomotion, with the parameter matrix (shown as a heatmap) appearing random due to initialization. By iteration~50, we observe significant structural changes in $\theta$, resulting in slow forward movement. Interestingly, from iteration 50 to 400, parameter updates become more incremental, suggesting a shift from exploration to exploitation. Despite these smaller parameter changes, the resulting behavior improves substantially: the agent discovers a high-speed locomotion strategy, propelling the swimmer out of view within approximately 200 time steps. This behavior illustrates the LLM’s capacity to refine policy parameters in a goal-directed manner, even under purely numerical supervision—effectively learning both coarse and fine-grained control strategies over time.
\section{\props: In-context Reasoning Example}\label{appendix:reasoning example}
\begin{figure}[h]
    \centering
    \includegraphics[width=\linewidth]{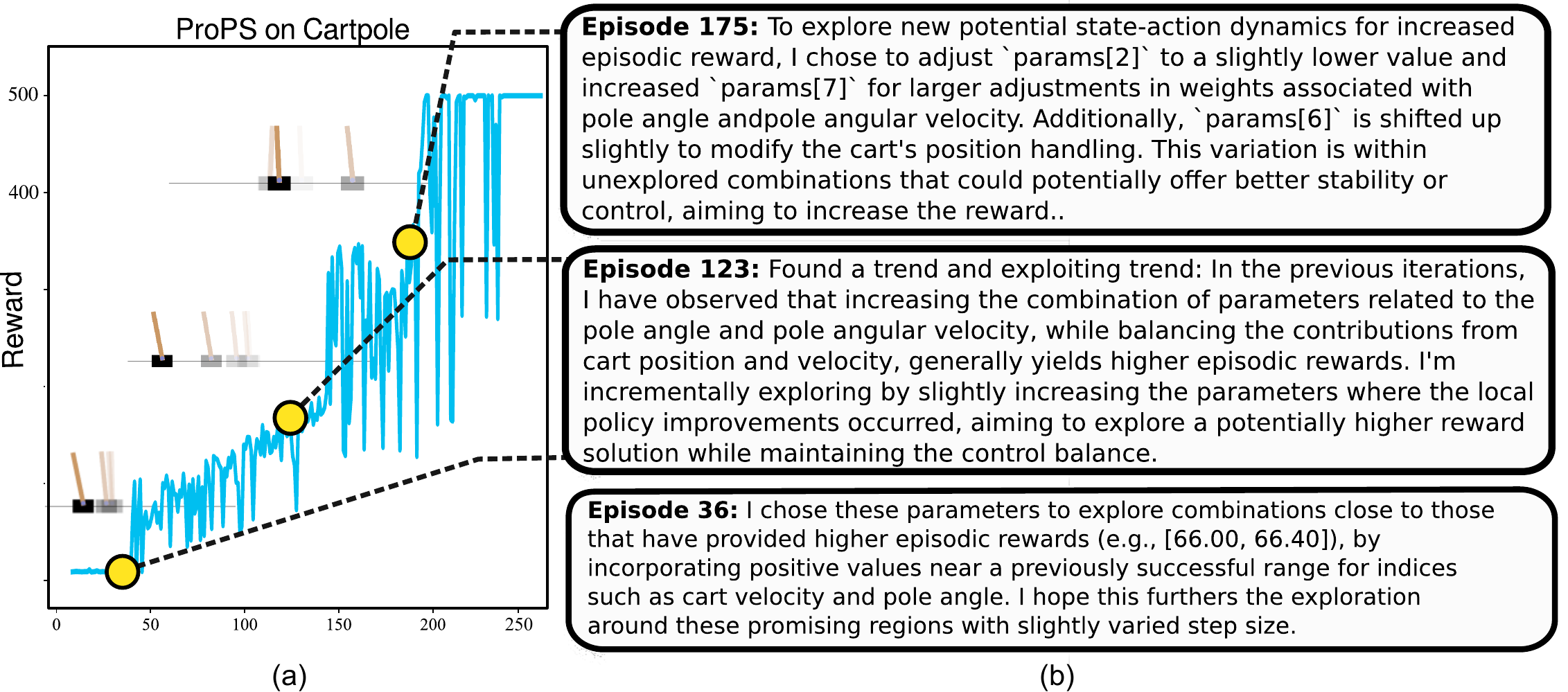}
    \caption{Summary of LLM's performance in Prompted Policy Search (\props) on the CartPole task: (a) demonstrates the ability of \props\ to iteratively reach optimal policies, and (b) highlights the LLMs capability in providing linguistic reasoning for the decisions made at each iteration in the policy search.}
    \label{fig:props_cartpole_text_example}
\end{figure}

In this section, we present an example of \props\ capabilities to perform policy search while providing textual justifications for the updates to policy parameters $\theta$ at each iteration. As presented in Fig.~\ref{fig:props_cartpole_text_example}, the underlying LLM performs well in terms of iteratively optimizing the episodic reward. Over the span of 250 iterations, the LLM is capable of consistently reaching policies that maximize the reward for the CartPole environment. This capability of the LLM is coupled with the provision of textual justification for the choice of updates for policy parameters at each iteration. These justifications, an example of which is highlighted in Fig.~\ref{fig:props_cartpole_text_example}(b), specify trends and patterns on how certain parameters or combinations of parameters impact the reward function, providing a level of interpretability and linguistic reasoning that is not afforded by traditional numeric optimizers.




\section{\propss\ with Hints: Ablation Study} \label{appendix:ablationstudy}
In this section, we investigate the relationship between the quality of the provided hint and its impact on \propss\ performance. To this end, we conduct an ablation study in four different domains: Mountain Car Continuous, Cliff Walking, Reacher, and Walker. Findings in this section demonstrate the role informative hints in enhancing \propss\ performance as an RL optimizer, marked by a more advantageous starting point, higher rewards, faster convergence, and reduced standard deviation.
\subsection{Mountain Car Continuous:}
We start by providing the following complete set of hints to \propss: 

\begin{tcolorbox}[colback=tbox_bg,colframe=tbox_frame,title=Expert Hint]
{\footnotesize
\textit{``(1) The force should be proportional to the velocity of the car. (2) When the velocity is negative, the force should be negative to push the car back. When the velocity is positive, the force should be positive to push the car forward. (3) The force should not be too high, as it penalizes the agent.''}}
\end{tcolorbox}

\clearpage
\begin{wrapfigure}{r}{0.48\textwidth}
\vspace{-12pt}
  \centering
  \includegraphics[trim = 5 18 5 8, clip, width=0.48\textwidth]{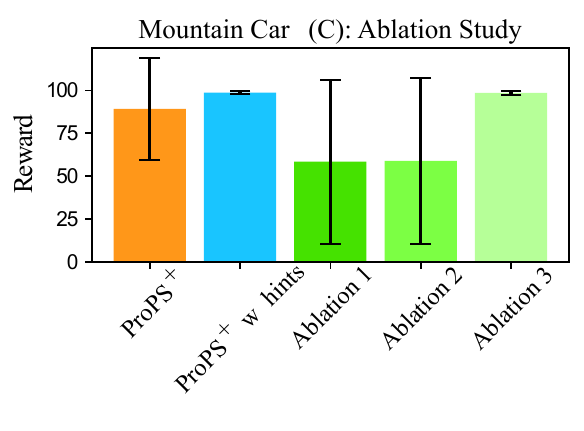}
  \caption{Performance of variations of hints, \propss\ vs \propss\ with hints: ablation study in the Mountain Car Continuous domain.}
  \label{fig:mc-ablation_study}
  \vspace{-15pt}
\end{wrapfigure}
The hint consists of three sections, in ablations 1, 2, and 3, we sequentially exclude the first, second, and third sections from the hint to investigate their effect, respectively. Results are shown in Fig.~\ref{fig:mc-ablation_study}. 

The ablation results reveal that omitting sentences with key instructions from the LLM prompt adversely affects performance, as reflected in the mean and increased standard deviation of maximum rewards in ablations 1 and 2. With the full hint included, the LLM consistently achieves the maximum reward, whereas, in ablations 1 and 2, where crucial instructions are omitted, only 50\% of trials reach the maximum reward. Conversely, removing the least informative hint segment allows the LLM to reach the maximum reward in 90\% of experiments in ablation 3. 
\subsection{Cliff Walking:}
The hint provided in this domain consists of two main components: instructions, and directional guidance linking the instructions to specific states and actions.
\begin{tcolorbox}[colback=tbox_bg,colframe=tbox_frame,title=Expert Hint: Cliff Walking]
{\footnotesize
\textit{``When the agent is on the top row (state = 0 to 11), the selected action should be moving right (action = 1). When the agent is on the second row (state = 12 to 23), the selected action should be moving right (action = 1). When the agent is on the third row (state = 24 to 35), the selected action should be moving up (action = 0). When the agent is at the starting point (state = 36), the selected action should be moving left (action = 3). When the agent is on the most right column (state = 11, 23, 35), the selected action should be moving down (action = 2). ''}}
\end{tcolorbox}

\begin{wrapfigure}{r}{0.48\textwidth}
\vspace{-12pt}
  \centering
  \includegraphics[trim = 5 10 5 6, clip, width=0.48\textwidth]{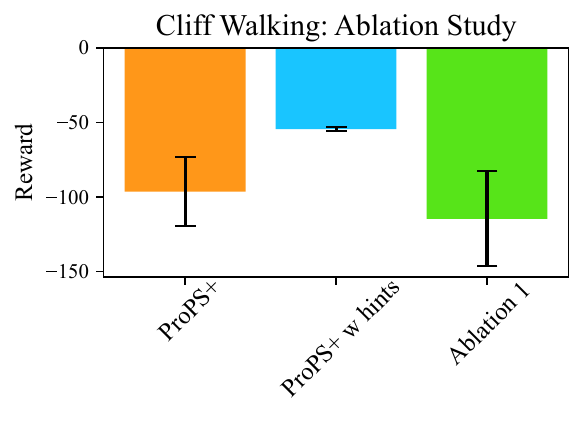}
  \caption{Performance of variations of hints, \propss\ vs \propss\ with hints: ablation study in the Cliff Walking domain.}
  \label{fig:CW-ablation_study}
  \vspace{-17pt}
\end{wrapfigure}
With the complete hint, \propss\ consistently achieves optimal policies, reaching a success rate of 100\% and often converging within the \textbf{first} iteration. To analyze the contribution of each component, we remove the directional guidance portion from all sentences in the hint. For instance, the first sentence in the reduced version becomes “When the agent is on the top row, the selected action should be moving right.”

As shown in Fig.~\ref{fig:CW-ablation_study}, removing the directional guidance (ablation~1) results in a substantial decline in performance, with the mean reward decreasing to -114.6, lower than that obtained by the original \propss\ prompt. The success rate similarly drops to 50\%, underscoring the critical role of detailed state-action associations in enabling the LLM to form robust and optimal policies.




\subsection{Reacher:}
The following hint is provided as a part of the prompt in the Reacher environment:
\begin{tcolorbox}[colback=tbox_bg,colframe=tbox_frame,title=Expert Hint: Reacher]
{\footnotesize
\textit{``(1) The policy should generate small, continuous torques that guide the fingertip directly toward the target. (2) Use shoulder joint torque for broad movements and elbow joint torque for fine adjustments. (3) As the fingertip nears the target, reduce torque to stabilize and minimize unnecessary motion.''}}
\end{tcolorbox}

\begin{wrapfigure}{r}{0.48\textwidth}
\vspace{-18pt}
  \centering
  \includegraphics[trim = 5 10 5 6, clip, width=0.48\textwidth]{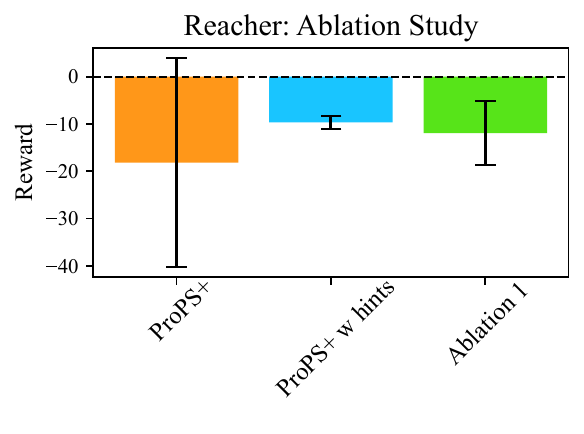}
  \caption{Performance of variations of hints, \propss\ vs \propss\ with hints: ablation study in the Reacher domain.}
  \label{fig:Reacher-ablation_study}
  \vspace{-18pt}
\end{wrapfigure}
Including the full hint in the prompt notably improves performance, with 70\% of experiments achieving rewards higher than -10, compared to only 30\% without hints. To assess the contribution of each component, we remove the second sentence of the hint, which provides crucial guidance on the coordination between the shoulder and elbow joints.

As illustrated in Fig.~\ref{fig:Reacher-ablation_study}, excluding this informative segment diminishes performance, reflected in a lower mean reward and an increase in the standard deviation of the best episodic rewards across training trials. Nevertheless, the reduced hint in ablation~1 still outperforms the baseline prompt without hints, indicating that even partial guidance can facilitate the RL policy search by biasing the exploration toward more promising action spaces.


\subsection{Walker:}
In this domain, we provide the following expert hint as part of the LLM prompt:
\begin{tcolorbox}[colback=tbox_bg,colframe=tbox_frame,title=Expert Hint: Walker]
{\footnotesize
\textit{``(1) Apply positive thigh torques (action[0] and action[3]) when the torso leans forward to drive motion, (2) coordinate leg and foot torques (action[1] and action[2], and action[4] and action[5]) to support ground contact and push-off, and alternate torque patterns between left and right limbs to maintain balance. (3) Modulate all torques smoothly using joint velocities to stabilize walking and avoid excessive, simultaneous activation.''}}
\end{tcolorbox}
\begin{wrapfigure}{r}{0.48\textwidth}
\vspace{-12pt}
\centering
\includegraphics[trim = 5 10 5 6, clip, width=0.48\textwidth]{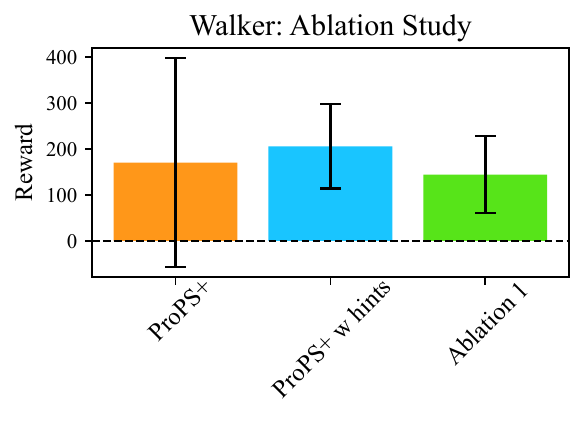}
\caption{Performance of variations of hints, \propss\ vs \propss\ with hints: ablation study in the Walker domain.}
\label{fig:Walker-ablation_study}
\vspace{-15pt}
\end{wrapfigure}
The hint for the Walker domain consists for three main components relating to: (1) driving motion forward, (2) coordinating between the different walker components, and (3) stabilizing the walker. Inclusion of the full hint improves policy quality and learning stability, with 60\% of runs achieving rewards above 120 points compared to only 20\% without hints. Despite the overall improvement, the initialization phase exhibited a higher variance due to random seed sensitivity, occasionally leading to slower convergence in certain trials.

To examine the influence of the coordination component, we remove the second sentence of the hint. As shown in Fig.~\ref{fig:Walker-ablation_study}, omitting this segment leads to a decline in performance, with mean rewards dropping to 144.06 and only 20\% of runs surpassing 120 points. These findings indicate that coordination-related guidance is essential for maintaining consistent gait dynamics and ensuring robust convergence across different runs.

\section{Details of the Experiment Tasks}
This section provides detailed descriptions of the environments used in our experiments. We included 11 tasks from the Gymnasium~\citep{towers2024gymnasium} and developed 4 custom tasks to evaluate the versatility and performance of our proposed method.
\subsection{Gymnasium Tasks}
\begin{enumerate}
    \item \textbf{FrozenLake-v1:}
    \begin{itemize}
        \item \textbf{Description:} The agent controls the movement of a character in a slippery grid world. Some tiles are walkable, and others lead to the agent falling into the water. The goal is to reach a reward tile. This is a discrete, grid-world environment.
        \item \textbf{Observation Space:} \texttt{Discrete(16)} - The agent's current position on the 4$\times$4 grid.
        \item \textbf{Action Space:} \texttt{Discrete(4)} - Move [Left, Down, Right, Up].
        \item \textbf{Reward:} +1 for reaching the goal tile, 0 otherwise.
        \item \textbf{Termination:} Agent reaches the goal, falls into a hole, or reaches 100 timesteps.
    \end{itemize}

    \item \textbf{CliffWalking-v0:}
    \begin{itemize}
        \item \textbf{Description:} The agent must navigate a 4$\times$12 slippery grid from a start state to a goal state. Falling off the cliff results in a large negative reward and resets the agent to the start.
        \item \textbf{Observation Space:} \texttt{Discrete(48)} - The agent's current position on the grid.
        \item \textbf{Action Space:} \texttt{Discrete(4)} - Move [Up, Right, Down, Left].
        \item \textbf{Reward:} -1 for each step, -100 for falling into the cliff. Reaching the goal provides 0 reward and ends the episode.
        \item \textbf{Termination:} Agent reaches the goal state or meets 100 timesteps.
    \end{itemize}

    \item \textbf{MountainCar-v0:}
    \begin{itemize}
        \item \textbf{Description:} A car is on a one-dimensional track, positioned between two "mountains". The goal is to drive up the mountain on the right; however, the car is not strong enough to scale the mountain in a single pass.
        \item \textbf{Observation Space:} \texttt{Box(2,)} - \texttt{[Car Position, Car Velocity]}.
        \item \textbf{Action Space:} \texttt{Discrete(3)} - Accelerate left, no acceleration, or accelerate right.
        \item \textbf{Reward:} -1 for each time step until the goal is reached.
        \item \textbf{Termination:} Car reaches the flag at position or the environment reaches 200 timesteps.
    \end{itemize}

    \item \textbf{CartPole-v1:}
    \begin{itemize}
        \item \textbf{Description:} A pole is attached by an un-actuated joint to a cart, which moves along a frictionless track. The system is controlled by applying a force of +1 or -1 to the cart. The pendulum starts upright, and the goal is to prevent it from falling over.
        \item \textbf{Observation Space:} \texttt{Box(4,)} - \texttt{[Cart Position, Cart Velocity, Pole Angle, Pole Angular Velocity]}.
        \item \textbf{Action Space:} \texttt{Discrete(2)} - Push cart to the left or right.
        \item \textbf{Reward:} +1 for every step taken.
        \item \textbf{Termination:} Pole angle $>$ 12 degrees, cart position $> \pm 2.4$, or episode length $>$ 500.
    \end{itemize}

    \item \textbf{MountainCarContinuous-v0:}
    \begin{itemize}
        \item \textbf{Description:} Similar to MountainCar, but with a continuous action space for applying force to the car.
        \item \textbf{Observation Space:} \texttt{Box(2,)} - \texttt{[Car Position, Car Velocity]}.
        \item \textbf{Action Space:} \texttt{Box(1,)} - Force applied to the car in $[-1.0, 1.0]$.
        \item \textbf{Reward:} +100 for reaching the goal, minus the sum of squares of actions taken ($0.1 \times \text{action}^2$).
        \item \textbf{Termination:} Car reaches the flag at position 0.45, or episode length $>$ 999.
    \end{itemize}

    \item \textbf{Swimmer-v5 (MuJoCo):}
    \begin{itemize}
        \item \textbf{Description:} A 2D three-link robot (swimmer) that must learn to swim forward by actuating its joints.
        \item \textbf{Observation Space:} \texttt{Box(8,)} - Angles and velocities of the swimmer's joints, as defined by MuJoCo.
        \item \textbf{Action Space:} \texttt{Box(2,)} - Torques applied to the 2 joints.
        \item \textbf{Reward:} Forward reward (difference in x-position) minus control cost.
        \item \textbf{Termination:} Episode length $>$ 1000.
    \end{itemize}

    \item \textbf{Hopper-v5 (MuJoCo):}
    \begin{itemize}
        \item \textbf{Description:} A 2D one-legged robot (hopper) that must learn to hop forward.
        \item \textbf{Observation Space:} \texttt{Box(11,)} - Positions and velocities of the hopper's body parts, as defined by MuJoCo.
        \item \textbf{Action Space:} \texttt{Box(3,)} - Torques applied to the 3 joints.
        \item \textbf{Reward:} Forward reward, alive bonus, minus control cost.
        \item \textbf{Termination:} Hopper falls (height $<$ 0.7m, angle $>$ 0.2 or $<$ -0.2 rad) or violates state constraints, or episode length $>$ 1000.
    \end{itemize}

    \item \textbf{InvertedPendulum-v5 (MuJoCo):}
    \begin{itemize}
        \item \textbf{Description:} A classic control problem where the goal is to balance a pendulum upright on a cart. 
        \item \textbf{Observation Space:} \texttt{Box(4,)} - \texttt{[cart\_x\_position, pole\_angle, cart\_x\_velocity, pole\_angular\_velocity]}.
        \item \textbf{Action Space:} \texttt{Box(1,)} - Force applied to the cart.
        \item \textbf{Reward:} +1 for every step the pendulum is upright.
        \item \textbf{Termination:} Pole falls over (absolute angle $>$ 0.2 radians), or episode length $>$ 1000.
    \end{itemize}

    \item \textbf{InvertedDoublePendulum-v5 (MuJoCo):}
    \begin{itemize}
        \item \textbf{Description:} A more challenging version of the inverted pendulum task with two linked poles. The goal is to balance both poles upright.
        \item \textbf{Observation Space:} \texttt{Box(11,)} - Positions, angles, and velocities of the cart and both poles.
        \item \textbf{Action Space:} \texttt{Box(1,)} - Force applied to the cart.
        \item \textbf{Reward:} Alive bonus minus distance penalty from upright minus velocity penalty.
        \item \textbf{Termination:} Lower pole tip goes below 1, or episode length $>$ 1000.
    \end{itemize}

    \item \textbf{Reacher-v5 (MuJoCo):}
    \begin{itemize}
        \item \textbf{Description:} A two-link robotic arm that must reach a randomly generated target location with its fingertip.
        \item \textbf{Observation Space:} \texttt{Box(11,)} - Arm angles, velocities, fingertip coordinates relative to target, and target coordinates.
        \item \textbf{Action Space:} \texttt{Box(2,)} - Torques applied to the 2 joints.
        \item \textbf{Reward:} Penalty of distance plus penalty of control.
        \item \textbf{Termination:} Episode length $>$ 50.
    \end{itemize}

    \item \textbf{Walker2d-v5 (MuJoCo):}
    \begin{itemize}
        \item \textbf{Description:} A 2D bipedal robot that must learn to walk forward.
        \item \textbf{Observation Space:} \texttt{Box(17,)} - Positions and velocities of the walker's body parts.
        \item \textbf{Action Space:} \texttt{Box(6,)} - Torques applied to the 6 joints.
        \item \textbf{Reward:} Forward reward plus alive bonus minus control cost.
        \item \textbf{Termination:} Walker falls (height $<$ 0.8 m or $>$ 2.0 m, absolute value of angle $>$ 1.0 rad) or episode length $>$ 1000.
    \end{itemize}
\end{enumerate}

\subsection{Customized Tasks}
\begin{enumerate}
    \item \textbf{Maze}~\citep{chan2020maze}
\begin{itemize}
    \item \textbf{Description:} The agent navigates a 2D $3 \times 3$ maze from a fixed start position to a fixed goal position. The maze contains impassable walls.
    \item \textbf{Observation Space:} \texttt{Discrete(9)} - the location of the agent in the maze.
    \item \textbf{Action Space:} \texttt{Discrete(4)} - Move [Up, Down, Right, Left].
    \item \textbf{Reward:} +1 for reaching the goal, -0.011 for each step taken.
    \item \textbf{Termination:} Agent reaches the goal or max episode steps (100) exceeded.
\end{itemize}

\item \textbf{Navigation}~\citep{germanis2024navigation}
\begin{itemize}
    \item \textbf{Description:} The agent must navigate a 2D track while avoiding hitting the track border. The agent has 5 fixed lidar sensors that returns the distances to the track border.
    \item \textbf{Observation Space:} \texttt{Box(5,)} representing the 5 measurements from the lidars.
    \item \textbf{Action Space:} \texttt{Discrete(3)} - Move forward, rotate left, and  rotate right.
    \item \textbf{Reward:} +5 for moving forward, -0.5 for rotating, and -200 for hitting the track border.
    \item \textbf{Termination:} The agents hits the tracker border or reaches 1000 timesteps.
\end{itemize}

\item \textbf{Nim}
\begin{itemize}
    \item \textbf{Description:} The game starts with 10 sticks. The agent and the environment take turns to remove 1 to 3 sticks. The player who takes the last stick loses. The environment has a rule-based optimal policy. The game starts with the agent taking the first move.
    \item \textbf{Observation Space:} \texttt{Box(11,)} - Integers representing the number of sticks left.
    \item \textbf{Action Space:} \texttt{Discrete(3)} - Number of sticks to remove.
    \item \textbf{Reward:} +1 for winning the game, -1 for losing the game.
    \item \textbf{Termination:} All sticks are removed.
\end{itemize}

\item \textbf{Pong}
\begin{itemize}
    \item \textbf{Description:} The agent controls a paddle on the left side, and a rule-based environment opponent controls the other one on the right side. The opponent is controlled by a rule-based optimal policy that will never miss the ball. The goal is to hit the ball as many times as possible.
    \item \textbf{Observation Space:} \texttt{Box(5,)} - ball position, velocity, and agent paddle's y position. 
    \item \textbf{Action Space:} \texttt{Discrete(3)} - Move paddle [Up, Stay, Down].
    \item \textbf{Reward:} +1 when the agent hits the ball.
    \item \textbf{Termination:} The agents misses the ball or reaches 3 hits.
\end{itemize}

\end{enumerate}

\section{Complete Prompts}\label{appendix:complete prompts}
\subsection{\props}
The introduced \props\ approach is a task-agnostic method that frames policy search as a numerical optimization problem. The objective is to identify policy parameters $\theta$ that maximize the expected mean episodic reward. Given our use of linear policies for continuous-state tasks and tabular policies for discrete-state tasks, the search space for $\theta$ varies accordingly. Consequently, the prompts provided to the Large Language Model (LLM) for parameter search are tailored respectively, though they maintain a consistent overarching structure.


Showing below is the prompt designed for a linear policy. It begins with an initial illustration of the optimization task. This is followed by a detailed description of the policy parameters and the corresponding search space. Subsequently, the prompt provides a sequential outline of the agentic steps the LLM is expected to perform during the search, concluding with a section of guidelines or ``tips'' that the LLM is instructed to adhere to throughout the optimization process.


\begin{tcolorbox}[enhanced, breakable, colback=tbox_bg,colframe=tbox_frame,title=\props\ Prompt for Linear Policies]
You are good global optimizer, helping me find the global maximum of a mathematical function f(params).
I will give you the function evaluation and the current iteration number at each step. 
Your goal is to propose input values that efficiently lead us to the global maximum within a limited number of iterations (400). \\
\\
\# Regarding the parameters **params**:
**params** is an array of {{ rank }} float numbers.
**params** values are in the range of [-6.0, 6.0] with 1 decimal place.
params represent a linear policy. 
f(params) is the episodic reward of the policy. \\

\# Here's how we'll interact: \\
1. I will first provide MAX\_STEPS (400) along with a few training examples. \\
2. You will provide your response in the following exact format: \\
    \hspace*{1cm}* Line 1: a new input 'params[0]: , params[1]: , params[2]: ,..., params[\promptvariable{\{\{ rank - 1 \\ \hspace*{1cm}\}\}}]: ', aiming to maximize the function's value f(params). \\
    \hspace*{1cm} Please propose params values in the range of [-6.0, 6.0], with 1 decimal place. \\
    \hspace*{1cm}* Line 2: detailed explanation of why you chose that input. \\
3. I will then provide the function's value f(params) at that point, and the current iteration. \\
4. We will repeat steps 2-3 until we reach the maximum number of iterations. \\

\# Remember: \\
1. **Do not propose previously seen params.** \\
2. **The global optimum should be around \promptvariable{\{\{ optimum \}\}}.** If you are below that, this is just a local optimum. You should explore instead of exploiting. \\
3. Search both positive and negative values. **During exploration, use search step size of \promptvariable{\{\{ step\_size \}\}}**.
\\
\\
Next, you will see examples of params and f(params) pairs. \\
\promptvariable{\{\{ episode\_reward\_buffer\_string \}\}} \\
\\
Now you are at iteration \promptvariable{\{\{step\_number\}\}} out of 400. Please provide the results in the indicated format. Do not provide any additional texts.
\end{tcolorbox}



For discrete-state tasks, tabular policies are employed. The corresponding full prompt, detailed subsequently, maintains the same fundamental structure as that for linear policies. The primary divergence is in the second paragraph, which describes the parameters. In this tabular context, each state is assigned a parameter that represents the index of the action to be selected upon observing that particular state, which can be found below:

\begin{tcolorbox}[enhanced, breakable, colback=tbox_bg,colframe=tbox_frame,title=\props\ Prompt for Tabular Policies]
You are good global optimizer, helping me find the global maximum of a mathematical function f(params).
I will give you the function evaluation and the current iteration number at each step. 
Your goal is to propose input values that efficiently lead us to the global maximum within a limited number of iterations (400). \\
\\
\# Regarding the parameters **params**:
**params** is an array of {{ rank }} int numbers.
**params** values should be an integer chosen from \promptvariable{\{\{ actions \}\}} \\

\# Here's how we'll interact: \\
1. I will first provide MAX\_STEPS (400) along with a few training examples. \\
2. You will provide your response in the following exact format: \\
    \hspace*{1cm}* Line 1: a new input 'params[0]: , params[1]: , params[2]: ,..., params[\promptvariable{\{\{ rank - 1 \\ \hspace*{1cm}\}\}}]: ', aiming to maximize the function's value f(params). \\
    \hspace*{1cm} Please propose params values from \promptvariable{\{\{ actions \}\}}. \\
    \hspace*{1cm}* Line 2: detailed explanation of why you chose that input. \\
3. I will then provide the function's value f(params) at that point, and the current iteration. \\
4. We will repeat steps 2-3 until we reach the maximum number of iterations. \\

\# Remember: \\
1. **Do not propose previously seen params.** \\
2. **The global optimum should be around \promptvariable{\{\{ optimum \}\}}.** If you are below that, this is just a local optimum. You should explore instead of exploiting. \\
3. Search all the possible values of params.
\\
\\
Next, you will see examples of params and f(params) pairs. \\
\promptvariable{\{\{ episode\_reward\_buffer\_string \}\}} \\
\\
Now you are at iteration \promptvariable{\{\{step\_number\}\}} out of 400. Please provide the results in the indicated format. Do not provide any additional texts.
\end{tcolorbox}

\clearpage
\subsection{\propss}

In \propss, we extend the basic \props\ by incorporating task-specific semantic contexts as another guidance for the policy search processes. To this end, we inserted another paragraph before the previous second paragraph, which describes the task environment details, as shown in the \propss\ prompt for linear policies below:

\begin{tcolorbox}[enhanced, breakable, colback=tbox_bg,colframe=tbox_frame,title=\propss\ Prompt for Linear Policies]
You are good global RL policy optimizer, helping me find the global optimal policy in the following environment: \\

\# Environment:
\promptvariable{\{\{ env\_description \}\}} \\

\# Regarding the parameters **params**:
**params** is an array of {{ rank }} float numbers.
**params** values are in the range of [-6.0, 6.0] with 1 decimal place.
params represent a linear policy. 
f(params) is the episodic reward of the policy. \\

\# Here's how we'll interact: \\
1. I will first provide MAX\_STEPS (400) along with a few training examples. \\
2. You will provide your response in the following exact format: \\
    \hspace*{1cm}* Line 1: a new input 'params[0]: , params[1]: , params[2]: ,..., params[\promptvariable{\{\{ rank - 1 \\ \hspace*{1cm}\}\}}]: ', aiming to maximize the function's value f(params). \\
    \hspace*{1cm} Please propose params values in the range of [-6.0, 6.0], with 1 decimal place. \\
    \hspace*{1cm}* Line 2: detailed explanation of why you chose that input. \\
3. I will then provide the function's value f(params) at that point, and the current iteration. \\
4. We will repeat steps 2-3 until we reach the maximum number of iterations. \\

\# Remember: \\
1. **Do not propose previously seen params.** \\
2. **The global optimum should be around \promptvariable{\{\{ optimum \}\}}.** If you are below that, this is just a local optimum. You should explore instead of exploiting. \\
3. Search both positive and negative values. **During exploration, use search step size of \promptvariable{\{\{ step\_size \}\}}**.
\\
\\
Next, you will see examples of params, and their episodic reward f(params). \\
\promptvariable{\{\{ episode\_reward\_buffer\_string \}\}} \\
\\
Now you are at iteration \promptvariable{\{\{step\_number\}\}} out of 400. Please provide the results in the indicated format. Do not provide any additional texts.
\end{tcolorbox}

The same applies to tabular policies. A paragraph for environment description is inserted before the original second paragraph. Please refer to our code base for detailed prompts.

\clearpage
The environment description includes what the domain is about, the observation space, the action space, the policy description, and the environment reward verbal descriptions. Below is an example for Inverted Pendulum task.

\begin{tcolorbox}[colback=tbox_bg,colframe=tbox_frame,title=\propss\ Prompt Environment Description for Inverted Pendulum]
The Inverted Pendulum environment consists of a cart that can be moved linearly, with a pole attached to one end and having another end free. The cart can be pushed left or right, and the goal is to balance the pole on top of the cart by applying forces to the cart. \\
The state is a vector of 4 elements, representing the cart position (-inf to inf m), pole angle (-inf to inf rad), cart velocity (-inf to inf m/s), and pole angular velocity (-inf to inf rad/s). The goal is to keep the pole upright and the cart within the bounding position of [-0.2, 0.2]. \\
The action space is 1 single float, representing the force to be applied on the cart (-3 to 3 N). \\
The policy is a linear policy with 5 parameters and works as follows:  \\
action = argmax(state @ W + b), where \\
state = [cart\_position, pole\_angle, cart\_velocity, pole\_angular\_velocity] \\
W = [[params[0]],
    
    [params[1]],
    
    [params[2]],
    
    [params[3]]]

b = [params[4]]

The reward is +1 for every time step the pole is upright. The episode ends when the pole falls over (the pole angle is larger than 0.2 radians or smaller than -0.2 radians).
\end{tcolorbox}

Below is the domain description for Pong. Note that the structure of the prompt is kept the same as in Inverted Pendulum above: The general task description, observation space, action space, reward, and policy descriptions.

\begin{tcolorbox}[colback=tbox_bg,colframe=tbox_frame,title=\propss\ Prompt Environment Description for Pong]
The pong environment is a 2D plane where the agent (robot) is a paddle on the left side (negative x side) and can move vertically freely (along y axis), and there is a ball moving in the plain. The goal of this environment is to always trying to hit the ball back. 

The state space consists of 5 dimensions: The paddle's y (state[0]); The ball's x (state[1]); The ball's y (state[2]); The ball's velocity on x (state[3]); The ball's velocity on y (state[4]). 

The action space consists of 3 actions (0: paddle move up, 1: paddle move down, 2: do nothing). 

Everytime the paddle hits the ball, the agent receives a reward of +1. If the paddle misses the ball, the game is over.

The policy is a linear policy with 18 parameters and works as follows: 

action = argmax(state @ W + B), where

state = [state[0], state[1], state[2], state[3], state[4]]

W = [[params[0], params[1]], [params[2]]

     [params[3]], [params[4], params[5]]
     
     [params[6]], [params[7], params[8]]
     
     [params[9]], [params[10], params[11]]
     
     [params[12]], [params[13], params[14]]
     
     ]

B = [params[15], params[16], params[17]]
\end{tcolorbox}

For more details, please refer to our code base.

\subsection{\propss\ with Hints}\label{appendix:propsswithhints}

The prompt utilized to provide the LLM with hints follows the same structure as \propss. The main difference is the inclusion of an entry for important hints, as illustrated below: 
\begin{tcolorbox}[colback=tbox_bg,colframe=tbox_frame,title=\propss\ Prompt with Hints]
{\footnotesize
You are a good global RL policy optimizer, helping me find an optimal policy in the following environment:\vspace{0.1cm}

\noindent
\begin{enumerate} [noitemsep,topsep=0pt, leftmargin=*]
    \item{\makebox[5cm][l]{Environment:} \textcolor{tbox_frame}{\% definition of the environment, parameters and policy}}
    \item{\makebox[5cm][l]{Regarding the parameters param:} \textcolor{tbox_frame}{\% definitions of parameters}}
    \item{\makebox[5cm][l]{Here's how we'll interact:} \textcolor{tbox_frame}{\% formatting instructions}}
    \item{\makebox[5cm][l]{Remember:} \textcolor{tbox_frame}{\% constraints to be respected}}
    \item{\makebox[5cm][l]{Important hints:} \textcolor{tbox_frame}{\% domain-specific expert hints}}
\end{enumerate}}
\end{tcolorbox} 

The following hints are provided for each of the three domains examined in this work:
\subsubsection{Mountain Car Continuous}
\begin{tcolorbox}[colback=tbox_bg,colframe=tbox_frame,title=Expert Hint: Mountain Car Continuous]
    {
    The optimal policy would follow the following behavior: The force should be proportional to the velocity of the car. When the velocity is negative, the force should be negative to push the car back. When the velocity is positive, the force should be positive to push the car forward. The force should not be too high, as it penalizes the agent.}
\end{tcolorbox}

\subsubsection{Navigation}
\begin{tcolorbox}[colback=tbox_bg,colframe=tbox_frame,title=Expert Hint: Navigation]
    {
    The optimal policy would follow the following behavior: When the robot is far from the wall, the agent should move forward (take action[0]). When the robot is close to the wall, the agent should rotate away from the wall, that is: if the sensors on the left show a shorter distance, the agent should rotate right (take action[2]), and if the sensors on the right show a shorter distance, the agent should rotate left (take action[1]).}
\end{tcolorbox}

\subsubsection{Inverted Double Pendulum}
\begin{tcolorbox}[colback=tbox_bg,colframe=tbox_frame,title=Expert Hint: Inverted Double Pendulum]
    {
    The optimal policy would follow the following behavior: The policy should generate smooth, low-magnitude forces to keep both poles upright and the cart centered. Correct angle and angular velocity deviations with proportional force. Use cart position and velocity to prevent drift. Avoid abrupt actions. When stable, minimize control effort to maintain balance and extend episode.}
\end{tcolorbox}

\subsection{Cliff Walking}
\begin{tcolorbox}[colback=tbox_bg,colframe=tbox_frame,title=Expert Hint: Cliff Walking]
    {
   The optimal policy would follow the following behavior: When the agent is on the top row (state = 0 to 11), the selected action should be moving right (action = 1). When the agent is on the second row (state = 12 to 23), the selected action should be moving right (action = 1). When the agent is on the third row (state = 24 to 35), the selected action should be moving up (action = 0). When the agent is at the starting point (state = 36), the selected action should be moving left (action = 3). When the agent is on the most right column (state = 11, 23, 35), the selected action should be moving down (action = 2).}
\end{tcolorbox}

\subsection{Reacher}
\begin{tcolorbox}[colback=tbox_bg,colframe=tbox_frame,title=Expert Hint: Reacher]
    {
    The optimal policy would follow the following behavior: The policy should generate small, continuous torques that guide the fingertip directly toward the target. Use shoulder joint torque for broad movements and elbow joint torque for fine adjustments. As the fingertip nears the target, reduce torque to stabilize and minimize unnecessary motion.}
\end{tcolorbox}

\subsection{Walker}
\begin{tcolorbox}[colback=tbox_bg,colframe=tbox_frame,title=Expert Hint: Walker]
    {
    The optimal policy would achieve the following behavior: apply positive thigh torques (action[0] and action[3]) when the torso leans forward to drive motion, coordinate leg and foot torques (action[1] and action[2], and action[4] and action[5]) to support ground contact and push-off, and alternate torque patterns between left and right limbs to maintain balance. It would modulate all torques smoothly using joint velocities to stabilize walking and avoid excessive, simultaneous activation.}
\end{tcolorbox}

\subsection{\props\ with Varied In-Context History Size}
In-context history are represented by placeholder \texttt{episode\_reward\_buffer\_string} in the prompts, which is composed of parameters with the evaluated rewards. Here is an example of the history:
\begin{tcolorbox}[colback=tbox_bg,colframe=tbox_frame,title=]
    {
params[0]: -2.1; params[1]: -1.6; params[2]: -2.6; f(params): -116.18 \\
params[0]: 4.1; params[1]: -2.4; params[2]: -4.4; f(params): -5611.77 \\
params[0]: 2.4; params[1]: 2.2; params[2]: -0.5; f(params): -571.42 \\
params[0]: -1.6; params[1]: -10.7; params[2]: -3.4; f(params): -447.54 \\
params[0]: -3.4; params[1]: 2; params[2]: 5.5; f(params): -4177.97 \\
params[0]: -6.3; params[1]: -4.5; params[2]: 1.2; f(params): -941.31
}
\end{tcolorbox}

When controlling the history size, we maintain a double-ended queue (deque) and remove the oldest entries whenever the history exceeds the predefined limit.

\section{Hyperparameters Used for Baselines}
We compared our proposed approach against seven widely adopted reinforcement learning algorithms: A2C, PPO, TRPO, DDPG, DQN, SAC, and TD3. Implementations for these algorithms were sourced from the publicly available Stable-Baselines3 and SB3-Contrib libraries~\citep{raffin2021stable}. For algorithms and tasks where hyperparameters were explicitly provided by these libraries, we directly adopted their recommended configurations. In cases where the libraries did not supply specific hyperparameters for a given task, we defaulted to the algorithms' standard hyperparameter settings. Representative examples of the hyperparameters used are detailed below, and the comprehensive configurations for all algorithms and tasks can be accessed in the associated configuration files within our codebase.

\begin{enumerate}
    \item \textbf{PPO on MountainCar-v0:}
    \begin{itemize}
        \item \textbf{normalize:} true
        \item \textbf{n\_envs:} 16
        \item \textbf{n\_steps:} 16
        \item \textbf{gae\_lambda:} 0.98
        \item \textbf{gamma:} 0.99
        \item \textbf{n\_epochs:} 4
        \item \textbf{ent\_coef:} 0.0
    \end{itemize}

    \item \textbf{TRPO on CartPole-v1:}
    \begin{itemize}
        \item \textbf{n\_envs:} 2
        \item \textbf{n\_steps:} 512
        \item \textbf{batch\_size:} 512
        \item \textbf{cg\_damping:} 1e-3
        \item \textbf{gae\_lambda:} 0.98
        \item \textbf{gamma:} 0.99
        \item \textbf{learning\_rate:} 1e-3
        \item \textbf{n\_critic\_updates:} 20
    \end{itemize}

    \item \textbf{SAC on MountainCarContinuous-v0:}
    \begin{itemize}
        \item \textbf{learning\_rate:} 3e-4
        \item \textbf{buffer\_sizes:} 50000
        \item \textbf{batch\_size:} 512
        \item \textbf{ent\_coef:} 0.1
        \item \textbf{train\_freq:} 32
        \item \textbf{gradient\_steps:} 32
        \item \textbf{gamma:} 0.9999
        \item \textbf{tau:} 0.01
        \item \textbf{learning\_starts:} 0
        \item \textbf{use\_sde:} True
    \end{itemize}
\end{enumerate}

For the rest of the algorithms and tasks, please refer to our codebase.


\newpage
\section*{NeurIPS Paper Checklist}

The checklist is designed to encourage best practices for responsible machine learning research, addressing issues of reproducibility, transparency, research ethics, and societal impact. Do not remove the checklist: {\bf The papers not including the checklist will be desk rejected.} The checklist should follow the references and follow the (optional) supplemental material.  The checklist does NOT count towards the page
limit. 

Please read the checklist guidelines carefully for information on how to answer these questions. For each question in the checklist:
\begin{itemize}
    \item You should answer \answerYes{}, \answerNo{}, or \answerNA{}.
    \item \answerNA{} means either that the question is Not Applicable for that particular paper or the relevant information is Not Available.
    \item Please provide a short (1–2 sentence) justification right after your answer (even for NA). 
\end{itemize}

{\bf The checklist answers are an integral part of your paper submission.} They are visible to the reviewers, area chairs, senior area chairs, and ethics reviewers. You will be asked to also include it (after eventual revisions) with the final version of your paper, and its final version will be published with the paper.

The reviewers of your paper will be asked to use the checklist as one of the factors in their evaluation. While "\answerYes{}" is generally preferable to "\answerNo{}", it is perfectly acceptable to answer "\answerNo{}" provided a proper justification is given (e.g., "error bars are not reported because it would be too computationally expensive" or "we were unable to find the license for the dataset we used"). In general, answering "\answerNo{}" or "\answerNA{}" is not grounds for rejection. While the questions are phrased in a binary way, we acknowledge that the true answer is often more nuanced, so please just use your best judgment and write a justification to elaborate. All supporting evidence can appear either in the main paper or the supplemental material, provided in appendix. If you answer \answerYes{} to a question, in the justification please point to the section(s) where related material for the question can be found.

IMPORTANT, please:
\begin{itemize}
    \item {\bf Delete this instruction block, but keep the section heading ``NeurIPS Paper Checklist"},
    \item  {\bf Keep the checklist subsection headings, questions/answers and guidelines below.}
    \item {\bf Do not modify the questions and only use the provided macros for your answers}.
\end{itemize}


\begin{enumerate}

\item {\bf Claims}
    \item[] Question: Do the main claims made in the abstract and introduction accurately reflect the paper's contributions and scope?
    \item[] Answer: \answerYes{} 
    \item[] Justification: The main claim in the abstract and introduction is that LLMs are capable of performing RL policy search, as the optimizer, combining both linguistic and numerical reasoning through in-context learning. We support this claim by evaluating the capabilities of a our systematic Prompted Policy Search (\props) approach in learning linear RL policies for 15 Gymnasium tasks.
    \item[] Guidelines:
    \begin{itemize}
        \item The answer NA means that the abstract and introduction do not include the claims made in the paper.
        \item The abstract and/or introduction should clearly state the claims made, including the contributions made in the paper and important assumptions and limitations. A No or NA answer to this question will not be perceived well by the reviewers. 
        \item The claims made should match theoretical and experimental results, and reflect how much the results can be expected to generalize to other settings. 
        \item It is fine to include aspirational goals as motivation as long as it is clear that these goals are not attained by the paper. 
    \end{itemize}

\item {\bf Limitations}
    \item[] Question: Does the paper discuss the limitations of the work performed by the authors?
    \item[] Answer: \answerYes{} 
    \item[] Justification: The limitations are discussed in Sec.~\ref{sec:discussion and limitations}.
    \item[] Guidelines:
    \begin{itemize}
        \item The answer NA means that the paper has no limitation while the answer No means that the paper has limitations, but those are not discussed in the paper. 
        \item The authors are encouraged to create a separate "Limitations" section in their paper.
        \item The paper should point out any strong assumptions and how robust the results are to violations of these assumptions (e.g., independence assumptions, noiseless settings, model well-specification, asymptotic approximations only holding locally). The authors should reflect on how these assumptions might be violated in practice and what the implications would be.
        \item The authors should reflect on the scope of the claims made, e.g., if the approach was only tested on a few datasets or with a few runs. In general, empirical results often depend on implicit assumptions, which should be articulated.
        \item The authors should reflect on the factors that influence the performance of the approach. For example, a facial recognition algorithm may perform poorly when image resolution is low or images are taken in low lighting. Or a speech-to-text system might not be used reliably to provide closed captions for online lectures because it fails to handle technical jargon.
        \item The authors should discuss the computational efficiency of the proposed algorithms and how they scale with dataset size.
        \item If applicable, the authors should discuss possible limitations of their approach to address problems of privacy and fairness.
        \item While the authors might fear that complete honesty about limitations might be used by reviewers as grounds for rejection, a worse outcome might be that reviewers discover limitations that aren't acknowledged in the paper. The authors should use their best judgment and recognize that individual actions in favor of transparency play an important role in developing norms that preserve the integrity of the community. Reviewers will be specifically instructed to not penalize honesty concerning limitations.
    \end{itemize}

\item {\bf Theory assumptions and proofs}
    \item[] Question: For each theoretical result, does the paper provide the full set of assumptions and a complete (and correct) proof?
    \item[] Answer: \answerNA{} 
    \item[] Justification: In this study, the assertions are backed by comprehensive empirical evidence. For each task investigated, we compare our method with seven leading RL algorithms across both continuous and discrete action spaces. To ensure fairness, all tasks and algorithms are assessed over 10 independent experiments.
    \item[] Guidelines:
    \begin{itemize}
        \item The answer NA means that the paper does not include theoretical results. 
        \item All the theorems, formulas, and proofs in the paper should be numbered and cross-referenced.
        \item All assumptions should be clearly stated or referenced in the statement of any theorems.
        \item The proofs can either appear in the main paper or the supplemental material, but if they appear in the supplemental material, the authors are encouraged to provide a short proof sketch to provide intuition. 
        \item Inversely, any informal proof provided in the core of the paper should be complemented by formal proofs provided in appendix or supplemental material.
        \item Theorems and Lemmas that the proof relies upon should be properly referenced. 
    \end{itemize}

    \item {\bf Experimental result reproducibility}
    \item[] Question: Does the paper fully disclose all the information needed to reproduce the main experimental results of the paper to the extent that it affects the main claims and/or conclusions of the paper (regardless of whether the code and data are provided or not)?
    \item[] Answer: \answerYes{} 
    \item[] Justification: The details of the experimental setup are described in Sec.~\ref{subsec:experimental setup}. Prompt structures for \props\ and \propss\ are presented in Figures~\ref{fig:promptProps}~and~\ref{fig:promptPropsPlus}. The examples for complete prompts, environment descriptions, and expert hints are provided in Appendix~\ref{appendix:complete prompts}. The full code base used to perform all experiments is provided as a part of the Supplementary material, all the is required is for the user to supply an API key for the LLM of choice (ChatGPT, Gemini, Claude). Additionally, we guarantee making the codebase publically available upon the acceptance of the paper. 
    \item[] Guidelines:
    \begin{itemize}
        \item The answer NA means that the paper does not include experiments.
        \item If the paper includes experiments, a No answer to this question will not be perceived well by the reviewers: Making the paper reproducible is important, regardless of whether the code and data are provided or not.
        \item If the contribution is a dataset and/or model, the authors should describe the steps taken to make their results reproducible or verifiable. 
        \item Depending on the contribution, reproducibility can be accomplished in various ways. For example, if the contribution is a novel architecture, describing the architecture fully might suffice, or if the contribution is a specific model and empirical evaluation, it may be necessary to either make it possible for others to replicate the model with the same dataset, or provide access to the model. In general. releasing code and data is often one good way to accomplish this, but reproducibility can also be provided via detailed instructions for how to replicate the results, access to a hosted model (e.g., in the case of a large language model), releasing of a model checkpoint, or other means that are appropriate to the research performed.
        \item While NeurIPS does not require releasing code, the conference does require all submissions to provide some reasonable avenue for reproducibility, which may depend on the nature of the contribution. For example
        \begin{enumerate}
            \item If the contribution is primarily a new algorithm, the paper should make it clear how to reproduce that algorithm.
            \item If the contribution is primarily a new model architecture, the paper should describe the architecture clearly and fully.
            \item If the contribution is a new model (e.g., a large language model), then there should either be a way to access this model for reproducing the results or a way to reproduce the model (e.g., with an open-source dataset or instructions for how to construct the dataset).
            \item We recognize that reproducibility may be tricky in some cases, in which case authors are welcome to describe the particular way they provide for reproducibility. In the case of closed-source models, it may be that access to the model is limited in some way (e.g., to registered users), but it should be possible for other researchers to have some path to reproducing or verifying the results.
        \end{enumerate}
    \end{itemize}

\item {\bf Open access to data and code}
    \item[] Question: Does the paper provide open access to the data and code, with sufficient instructions to faithfully reproduce the main experimental results, as described in supplemental material?
    \item[] Answer: \answerYes{} 
    \item[] Justification: We are submitting the code for review. We will provide public access to the code after the paper is accepted. 
    \item[] Guidelines:
    \begin{itemize}
        \item The answer NA means that paper does not include experiments requiring code.
        \item Please see the NeurIPS code and data submission guidelines (\url{https://nips.cc/public/guides/CodeSubmissionPolicy}) for more details.
        \item While we encourage the release of code and data, we understand that this might not be possible, so “No” is an acceptable answer. Papers cannot be rejected simply for not including code, unless this is central to the contribution (e.g., for a new open-source benchmark).
        \item The instructions should contain the exact command and environment needed to run to reproduce the results. See the NeurIPS code and data submission guidelines (\url{https://nips.cc/public/guides/CodeSubmissionPolicy}) for more details.
        \item The authors should provide instructions on data access and preparation, including how to access the raw data, preprocessed data, intermediate data, and generated data, etc.
        \item The authors should provide scripts to reproduce all experimental results for the new proposed method and baselines. If only a subset of experiments are reproducible, they should state which ones are omitted from the script and why.
        \item At submission time, to preserve anonymity, the authors should release anonymized versions (if applicable).
        \item Providing as much information as possible in supplemental material (appended to the paper) is recommended, but including URLs to data and code is permitted.
    \end{itemize}

\item {\bf Experimental setting/details}
    \item[] Question: Does the paper specify all the training and test details (e.g., data splits, hyperparameters, how they were chosen, type of optimizer, etc.) necessary to understand the results?
    \item[] Answer: \answerYes{} 
    \item[] Justification: We provide all experimental details needed for reproducibility. Although the paper does not explicitly detail all hyperparameters for the seven baseline RL algorithms (for the sake of brevity), we reference the publically maintained Stable-Baselines3 and SB3-Contrib libraries \citep{raffin2021stable}, which we utilize in this work.
    \item[] Guidelines:
    \begin{itemize}
        \item The answer NA means that the paper does not include experiments.
        \item The experimental setting should be presented in the core of the paper to a level of detail that is necessary to appreciate the results and make sense of them.
        \item The full details can be provided either with the code, in appendix, or as supplemental material.
    \end{itemize}

\item {\bf Experiment statistical significance}
    \item[] Question: Does the paper report error bars suitably and correctly defined or other appropriate information about the statistical significance of the experiments?
    \item[] Answer: \answerYes{} 
    \item[] Justification: We report the standard deviation in our results for all tasks and algorithms. This can be seen in all tables and figures where this information is relevant in the paper. The only exception is in Figures~\ref{fig:times}~and~\ref{fig:swimmer-continuous-tasks}, in which the standard deviations are excluded for a cleaner visual representation.
    \item[] Guidelines:
    \begin{itemize}
        \item The answer NA means that the paper does not include experiments.
        \item The authors should answer "Yes" if the results are accompanied by error bars, confidence intervals, or statistical significance tests, at least for the experiments that support the main claims of the paper.
        \item The factors of variability that the error bars are capturing should be clearly stated (for example, train/test split, initialization, random drawing of some parameter, or overall run with given experimental conditions).
        \item The method for calculating the error bars should be explained (closed form formula, call to a library function, bootstrap, etc.)
        \item The assumptions made should be given (e.g., Normally distributed errors).
        \item It should be clear whether the error bar is the standard deviation or the standard error of the mean.
        \item It is OK to report 1-sigma error bars, but one should state it. The authors should preferably report a 2-sigma error bar than state that they have a 96\% CI, if the hypothesis of Normality of errors is not verified.
        \item For asymmetric distributions, the authors should be careful not to show in tables or figures symmetric error bars that would yield results that are out of range (e.g. negative error rates).
        \item If error bars are reported in tables or plots, The authors should explain in the text how they were calculated and reference the corresponding figures or tables in the text.
    \end{itemize}

\item {\bf Experiments compute resources}
    \item[] Question: For each experiment, does the paper provide sufficient information on the computer resources (type of compute workers, memory, time of execution) needed to reproduce the experiments?
    \item[] Answer: \answerYes{} 
    \item[] Justification: All of the experiments utilize LLMs as the optimizer and compare them to baseline algorithms. We describe the compute resources used in this work, in Sec.~\ref{subsec:experimental setup}.
    \item[] Guidelines:
    \begin{itemize}
        \item The answer NA means that the paper does not include experiments.
        \item The paper should indicate the type of compute workers CPU or GPU, internal cluster, or cloud provider, including relevant memory and storage.
        \item The paper should provide the amount of compute required for each of the individual experimental runs as well as estimate the total compute. 
        \item The paper should disclose whether the full research project required more compute than the experiments reported in the paper (e.g., preliminary or failed experiments that didn't make it into the paper). 
    \end{itemize}
    
\item {\bf Code of ethics}
    \item[] Question: Does the research conducted in the paper conform, in every respect, with the NeurIPS Code of Ethics \url{https://neurips.cc/public/EthicsGuidelines}?
    \item[] Answer: \answerYes{} 
    \item[] Justification: The contributions of this paper focus on the capability of LLMs to combine numerical and semantic reasoning in optimization problems, more specifically RL policy search. The evidence we present, including fine-tuning of small LLM models, do not have direct safety or security implications.
    \item[] Guidelines:
    \begin{itemize}
        \item The answer NA means that the authors have not reviewed the NeurIPS Code of Ethics.
        \item If the authors answer No, they should explain the special circumstances that require a deviation from the Code of Ethics.
        \item The authors should make sure to preserve anonymity (e.g., if there is a special consideration due to laws or regulations in their jurisdiction).
    \end{itemize}

\item {\bf Broader impacts}
    \item[] Question: Does the paper discuss both potential positive societal impacts and negative societal impacts of the work performed?
    \item[] Answer: \answerYes{} 
    \item[] Justification: In Sec.~\ref{sec:discussion and limitations} we discuss broader impacts.
    \item[] Guidelines:
    \begin{itemize}
        \item The answer NA means that there is no societal impact of the work performed.
        \item If the authors answer NA or No, they should explain why their work has no societal impact or why the paper does not address societal impact.
        \item Examples of negative societal impacts include potential malicious or unintended uses (e.g., disinformation, generating fake profiles, surveillance), fairness considerations (e.g., deployment of technologies that could make decisions that unfairly impact specific groups), privacy considerations, and security considerations.
        \item The conference expects that many papers will be foundational research and not tied to particular applications, let alone deployments. However, if there is a direct path to any negative applications, the authors should point it out. For example, it is legitimate to point out that an improvement in the quality of generative models could be used to generate deepfakes for disinformation. On the other hand, it is not needed to point out that a generic algorithm for optimizing neural networks could enable people to train models that generate Deepfakes faster.
        \item The authors should consider possible harms that could arise when the technology is being used as intended and functioning correctly, harms that could arise when the technology is being used as intended but gives incorrect results, and harms following from (intentional or unintentional) misuse of the technology.
        \item If there are negative societal impacts, the authors could also discuss possible mitigation strategies (e.g., gated release of models, providing defenses in addition to attacks, mechanisms for monitoring misuse, mechanisms to monitor how a system learns from feedback over time, improving the efficiency and accessibility of ML).
    \end{itemize}
    
\item {\bf Safeguards}
    \item[] Question: Does the paper describe safeguards that have been put in place for responsible release of data or models that have a high risk for misuse (e.g., pretrained language models, image generators, or scraped datasets)?
    \item[] Answer: \answerNA{} 
    \item[] Justification: We utilize publicly available models and benchmark simulation tasks, all of which are already in the public domain. Therefore, the risk of misuse is minimal, as comparable reinforcement learning environments are readily accessible from various open-source projects, such as Gymnasium~\citep{towers2024gymnasium}.
    

    \item[] Guidelines:
    \begin{itemize}
        \item The answer NA means that the paper poses no such risks.
        \item Released models that have a high risk for misuse or dual-use should be released with necessary safeguards to allow for controlled use of the model, for example by requiring that users adhere to usage guidelines or restrictions to access the model or implementing safety filters. 
        \item Datasets that have been scraped from the Internet could pose safety risks. The authors should describe how they avoided releasing unsafe images.
        \item We recognize that providing effective safeguards is challenging, and many papers do not require this, but we encourage authors to take this into account and make a best faith effort.
    \end{itemize}

\item {\bf Licenses for existing assets}
    \item[] Question: Are the creators or original owners of assets (e.g., code, data, models), used in the paper, properly credited and are the license and terms of use explicitly mentioned and properly respected?
    \item[] Answer: \answerYes{} 
    \item[] Justification: We cite all the resources used in our evaluations, including the algorithms used for the baselines \citep{raffin2021stable}, the Gymnasium environments \citep{towers2024gymnasium, todorov2012mujoco}, and LLMs at the center of the optimization \citep{anthropic_claude_3_7_2025, openai_chatgpt, team2023gemini}.
    \item[] Guidelines:
    \begin{itemize}
        \item The answer NA means that the paper does not use existing assets.
        \item The authors should cite the original paper that produced the code package or dataset.
        \item The authors should state which version of the asset is used and, if possible, include a URL.
        \item The name of the license (e.g., CC-BY 4.0) should be included for each asset.
        \item For scraped data from a particular source (e.g., website), the copyright and terms of service of that source should be provided.
        \item If assets are released, the license, copyright information, and terms of use in the package should be provided. For popular datasets, \url{paperswithcode.com/datasets} has curated licenses for some datasets. Their licensing guide can help determine the license of a dataset.
        \item For existing datasets that are re-packaged, both the original license and the license of the derived asset (if it has changed) should be provided.
        \item If this information is not available online, the authors are encouraged to reach out to the asset's creators.
    \end{itemize}

\item {\bf New assets}
    \item[] Question: Are new assets introduced in the paper well documented and is the documentation provided alongside the assets?
    \item[] Answer: \answerYes{} 
    \item[] Justification: When the code is released, the proper documentation standards will be followed.
    \item[] Guidelines:
    \begin{itemize}
        \item The answer NA means that the paper does not release new assets.
        \item Researchers should communicate the details of the dataset/code/model as part of their submissions via structured templates. This includes details about training, license, limitations, etc. 
        \item The paper should discuss whether and how consent was obtained from people whose asset is used.
        \item At submission time, remember to anonymize your assets (if applicable). You can either create an anonymized URL or include an anonymized zip file.
    \end{itemize}

\item {\bf Crowdsourcing and research with human subjects}
    \item[] Question: For crowdsourcing experiments and research with human subjects, does the paper include the full text of instructions given to participants and screenshots, if applicable, as well as details about compensation (if any)? 
    \item[] Answer: \answerNA{} 
    \item[] Justification: In this work, no crowd-sourcing or research involving human subjects is included.
    \item[] Guidelines:
    \begin{itemize}
        \item The answer NA means that the paper does not involve crowdsourcing nor research with human subjects.
        \item Including this information in the supplemental material is fine, but if the main contribution of the paper involves human subjects, then as much detail as possible should be included in the main paper. 
        \item According to the NeurIPS Code of Ethics, workers involved in data collection, curation, or other labor should be paid at least the minimum wage in the country of the data collector. 
    \end{itemize}

\item {\bf Institutional review board (IRB) approvals or equivalent for research with human subjects}
    \item[] Question: Does the paper describe potential risks incurred by study participants, whether such risks were disclosed to the subjects, and whether Institutional Review Board (IRB) approvals (or an equivalent approval/review based on the requirements of your country or institution) were obtained?
    \item[] Answer: \answerNA{} 
    \item[] Justification: This paper does not involve crowd-sourcing or research with human subjects.
    \item[] Guidelines:
    \begin{itemize}
        \item The answer NA means that the paper does not involve crowdsourcing nor research with human subjects.
        \item Depending on the country in which research is conducted, IRB approval (or equivalent) may be required for any human subjects research. If you obtained IRB approval, you should clearly state this in the paper. 
        \item We recognize that the procedures for this may vary significantly between institutions and locations, and we expect authors to adhere to the NeurIPS Code of Ethics and the guidelines for their institution. 
        \item For initial submissions, do not include any information that would break anonymity (if applicable), such as the institution conducting the review.
    \end{itemize}

\item {\bf Declaration of LLM usage}
    \item[] Question: Does the paper describe the usage of LLMs if it is an important, original, or non-standard component of the core methods in this research? Note that if the LLM is used only for writing, editing, or formatting purposes and does not impact the core methodology, scientific rigorousness, or originality of the research, declaration is not required.
    \item[] Answer: \answerYes{} 
    \item[] Justification: This paper specifically investigates the ability of LLMs to perform RL policy search in a novel approach that combines numerical and linguistic reasoning.
    \item[] Guidelines:
    \begin{itemize}
        \item The answer NA means that the core method development in this research does not involve LLMs as any important, original, or non-standard components.
        \item Please refer to our LLM policy (\url{https://neurips.cc/Conferences/2025/LLM}) for what should or should not be described.
    \end{itemize}

\end{enumerate}

\stopcontents[apx] 
\end{document}